\documentclass{article} % For LaTeX2e
\usepackage{iclr2025_conference,times}

\usepackage{amsmath,amsfonts,bm}

\def\eqref#1{equation~\ref{#1}}
\def\1{\bm{1}}

\DeclareMathAlphabet{\mathsfit}{\encodingdefault}{\sfdefault}{m}{sl}
\SetMathAlphabet{\mathsfit}{bold}{\encodingdefault}{\sfdefault}{bx}{n}

\usepackage{makecell}
\usepackage{float}
\usepackage{graphicx}
\usepackage[utf8]{inputenc} % allow utf-8 input
\usepackage[T1]{fontenc}    % use 8-bit T1 fonts
\usepackage{hyperref}       % hyperlinks
\usepackage{url}            % simple URL typesetting
\usepackage{booktabs}       % professional-quality tables
\usepackage{amsfonts}       % blackboard math symbols
\usepackage{nicefrac}       % compact symbols for 1/2, etc.
\usepackage{microtype}      % microtypography
\usepackage{xcolor}         % colors
\usepackage{amsmath}
\usepackage{amssymb}
\usepackage{bbm}
\usepackage[ruled]{algorithm2e}
\usepackage{algpseudocode}
\usepackage{multirow}
\usepackage{array}
\usepackage{siunitx}
\usepackage{caption}
\hypersetup{
colorlinks = true,
citecolor = {blue},
urlcolor = {cyan}
}

\newcommand{\model}{LLM-Streamline}
\newcommand{\smodel}{LLM-Streamline }

\title{Streamlining Redundant Layers to Compress Large Language Models}

\author{
	Xiaodong Chen$^{1}$\footnotemark[2], Yuxuan Hu$^{1}$\footnotemark[2], Jing Zhang$^{1}$\footnotemark[1], Yanling Wang$^{2}$, Cuiping Li$^{1}$, Hong Chen$^{1}$\\ 
	$^1$ Renmin University of China, China\\
        $^2$ Zhongguancun Laboratory, China\\
	\texttt{chenxiaodong@ruc.edu.cn} \\
	\texttt{\{huyuxuan1999,zhang-jing,licuiping,chong\}@ruc.edu.cn} \\
        \texttt{wangyl@zgclab.edu.cn} \\
}
 
 \footnotetext[2]{These authors contributed equally.}
 \footnotetext[1]{Jing Zhang is the corresponding author.}

\begin{document}

\maketitle

\begin{abstract}
This paper introduces LLM-Streamline, a pioneer work on layer pruning for large language models (LLMs). It is based on the observation that different layers have varying impacts on hidden states, enabling the identification of less important layers to be pruned. 
LLM-Streamline comprises two parts: layer pruning, which removes consecutive layers with the lowest importance based on target sparsity, and layer replacement, a novel module that trains a lightweight network to replace the pruned layers to mitigate performance loss. Additionally, a new metric called stability is proposed to address the limitations of the widely used accuracy metric in evaluating model compression. Experiments show that LLM-Streamline outperforms both previous and concurrent state-of-the-art pruning methods in terms of both performance and training efficiency. Our code is available at \href{https://github.com/RUCKBReasoning/LLM-Streamline}{this repository}.

\end{abstract}

\begin{figure}[h] % h表示here，t表示top
  \centering % 使图片居中
  \includegraphics[width=0.85\textwidth]{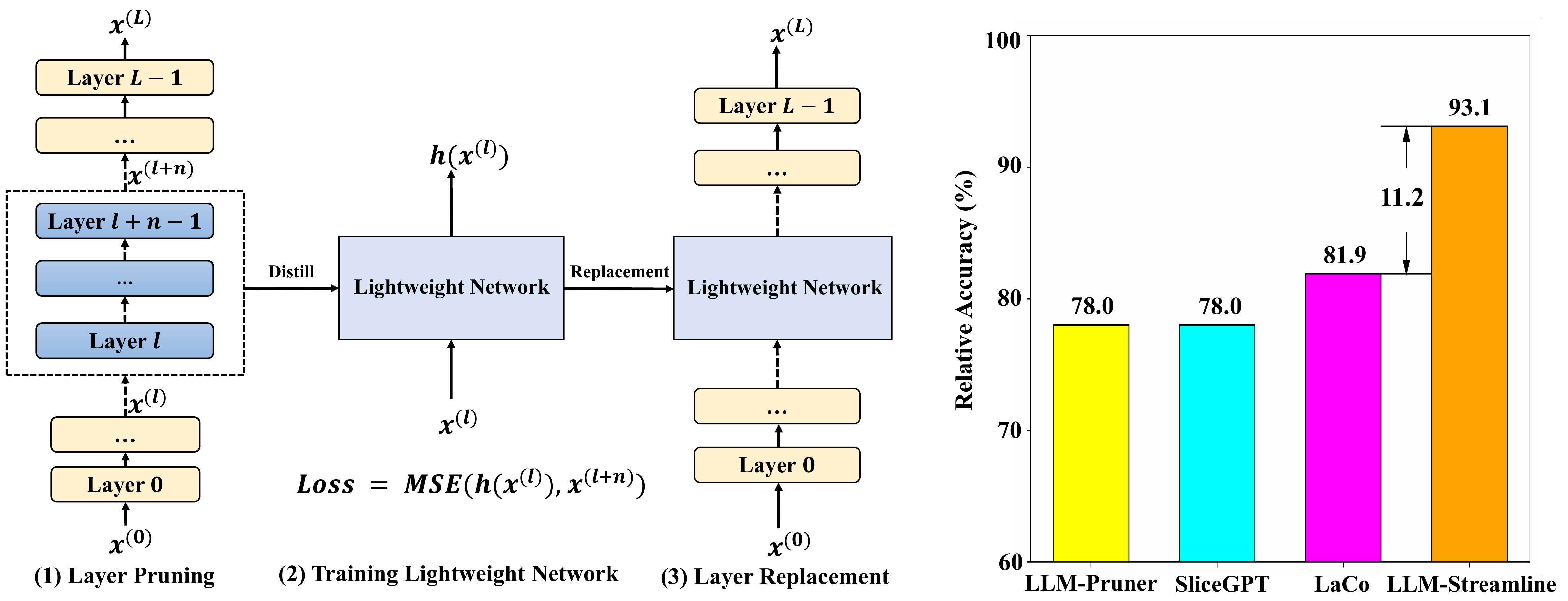} % 插入图片，设置宽度为文本宽度的50%
  \caption{The left side of the figure illustrates the LLM-Streamline workflow, which includes layer pruning to remove consecutive layers and layer replacement where a lightweight network is trained to replace the pruned layers. The right side of the figure presents the comparison results of LLM-Streamline with the state-of-the-art (SOTA) pruning methods on 12 classification benchmarks (details in Section~\ref{sec: benchmark}) after pruning about 25\% of the parameters on Llama2-7B. LLM-Streamline achieves 11.2\% higher relative accuracy than these methods, where the relative accuracy represents the
percentage of the original model’s accuracy retained by the pruning method.} % 添加标题

  \label{fig: workflow} % 定义标签
\end{figure}

\section{Introduction\label{sec: introduction}}

Large language models (LLMs) built on the Transformer architecture~\citep{vaswani2017attention} have gained widespread attention and are applied across diverse domains and tasks. However, as LLMs increase in size, their hardware requirements escalate substantially, thereby constraining their applicability and impeding their deployment in real-world scenarios.
To reduce the hardware requirements for deploying LLMs, research efforts have been devoted to developing compact models that maintain high performance through model compression~\citep{zhu2023survey, wang2024survey}. 
% These compact models significantly reduce memory consumption, computation time, and deployment costs.
Currently, model compression techniques are widely categorized into knowledge distillation~\citep{hinton2015distilling, gou2021knowledge, li2022apidistill, huang2022distill, ho2022distill}, quantization~\citep{liu2021quan, gholami2022survey, dettmers2022int8}, and pruning~\citep{louizos2017l0, chen2023lorashear, frantar2023sparsegpt, das2023beyond, sun2023simple, xia2023sheared}. Knowledge distillation achieves compression by transferring the capabilities of a larger teacher model to a smaller student model. Quantization compresses the model by quantizing the weights to lower precision. Alternatively, pruning compresses the model by eliminating unimportant parameters and modules. 

This work focuses on the popular pruning methods.
Previous approaches for LLM primarily prune dense matrices~\citep{ashkboos2024slicegpt}, attention heads~\citep{michel2019sixteen, voita2019mha}, filters~\citep{mccarley2019filter, prasanna2020filter}, or prune parameters to reduce an LLM's hidden dimension~\citep{xia2023sheared, van2023llm, sp3-hu2024}.
Despite their effectiveness, these methods often result in structural irregularities, making it inflexible to store and deploy the pruned models. 
In contrast, layer pruning method simply reduces the depth of LLMs.
As the layers of LLMs are stored in data structures like nn.ModuleList, layer pruning only requires removing elements from this list, making it more flexible for application. Therefore, exploring an effective layer-wise pruning method is crucial.

\begin{figure}[t] % h表示here，t表示top
  \centering % 使图片居中
  \includegraphics[width=0.6\textwidth]{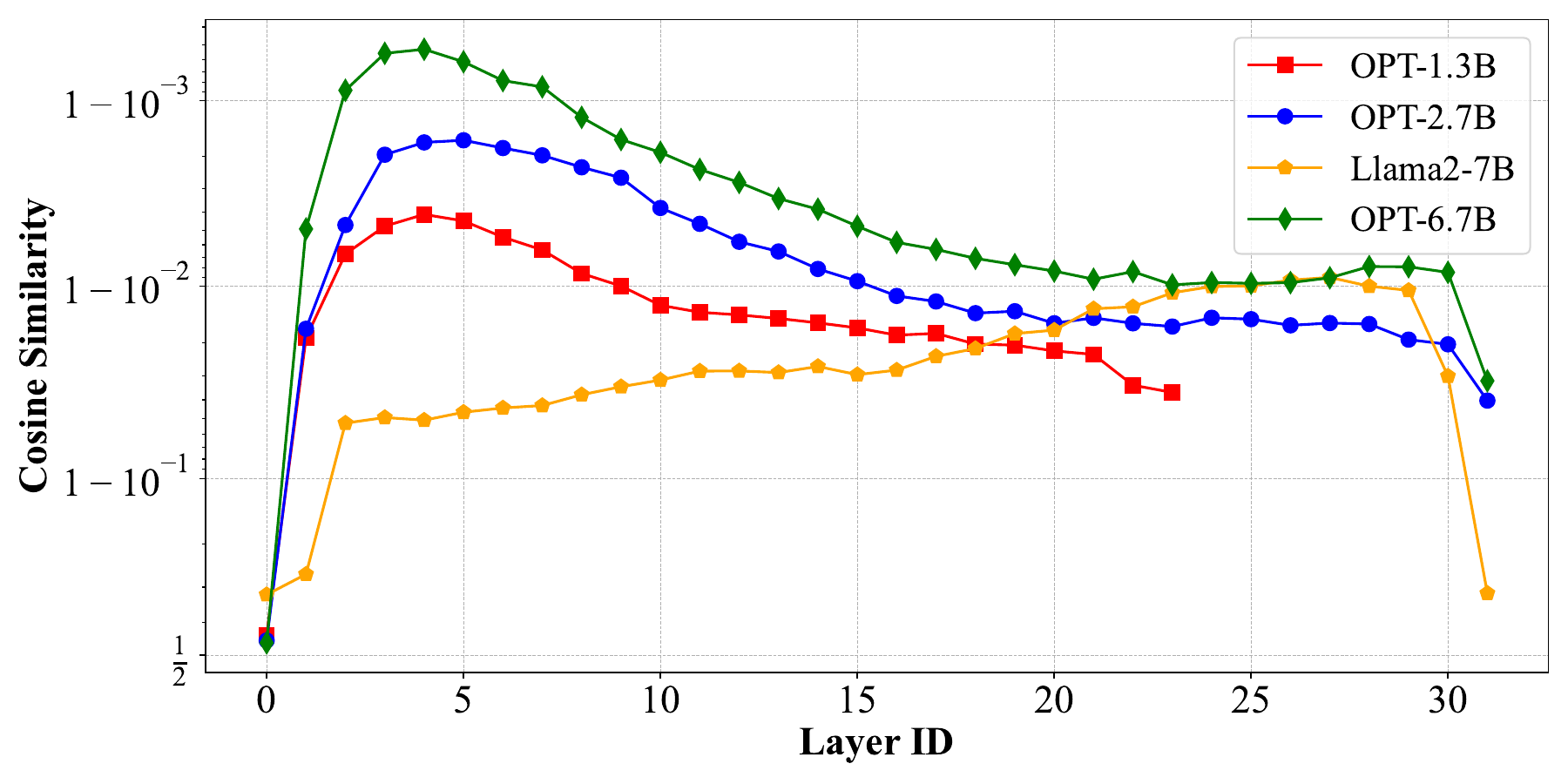} % 插入图片，设置宽度为文本宽度的50%
  \caption{The cosine similarity between the input and output hidden states of each layer in OPT-1.3B, OPT-2.7B, OPT-6.7B, and Llama2-7B.} % 添加标题
  \label{fig:cosine_similarity} % 定义标签
\end{figure}

The core idea of layer pruning is to identify and remove less important layers in an LLM. Specifically, the effect of each layer can be viewed as a transformation of the hidden states. If the input and output hidden states of a particular layer are highly similar, such as exhibiting high cosine similarity, we can say that the layer has a small impact on adjusting the hidden states.
As illustrated in Fig.~\ref{fig:cosine_similarity}, our pilot study shows that certain contiguous layers indeed have smaller impact on the hidden states, indicating they are less important and suitable for pruning. 
Some concurrent works~\citep{song2024slebstreamliningllmsredundancy, kim2024shortened, yang2024laco, men2024shortgpt, gromov2024unreasonable} also explore layer pruning.
%\footnote{These works are publicly available as preprints.} 
These studies either prune unimportant layers directly without further training~\citep{song2024slebstreamliningllmsredundancy, men2024shortgpt} or fine-tune the pruned model to enhance performance~\citep{kim2024shortened, yang2024laco, gromov2024unreasonable}. 
Directly removing layers can lead to more performance degradation.
While parameter-efficient fine-tuning techniques like LoRA~\citep{hu2021lora} are used to train the pruned LLM, fine-tuning the model to make the original non-contiguous layers compensate for the performance degradation is not an easy task (details in Section~\ref{sec: layer replacement}).

In this work, we propose a layer pruning method called \model, which exhibits advantages in both performance and training efficiency while requiring less training data. \smodel comprises two components: layer pruning and layer replacement.
According to a certain target sparsity, the first step removes consecutive layers with the lowest importance from the original LLM. Subsequently, we train a lightweight network to replace the pruned layers, aiming to recover the performance degradation caused by pruning. We can employ various architectures for this lightweight  network, including a feed-forward neural network (FFN), a SwiGLU-based feed-forward neural network (SwiGLU), and a Transformer layer.

Additionally, we find that existing accuracy metrics for evaluating model compression methods have limitations. Specifically, in natural language understanding tasks that involve multiple-choice classification, a compressed model may guess correct answers for samples on which the original model was uncertain, resulting in an overestimation of the compression performance. To address this issue, we propose a new metric named stability, which measures the consistency of predictions before and after pruning, considering the prediction confidence of the original model.

Overall, this paper makes the following contributions:

\begin{itemize}
    \item We propose \model, a layer-wise pruning algorithm that demonstrates superior effectiveness and efficiency compared to concurrent methods. To mitigate the potential performance degradation caused by pruning, we propose to use a lightweight network to approximate the functionality of the pruned layers.
    \item We propose a new metric called stability, which considers both the prediction confidence of the original model and the consistency of predictions before and after pruning. Stability provides a more accurate reflection of the pruned model's performance in classification tasks compared to the widely used accuracy metric.
    \item We conduct experiments on 12 well-known classification benchmarks and 3 generation benchmarks. Our results show that for an LLM with 7B or 13B parameters and a 25\% pruning rate, we can maintain 93\% performance in classification tasks and 77\% in generation tasks without requiring a lot of training data, outperforming existing SOTA pruning methods.
\end{itemize}

% \textbf{Note:} As we were finalizing this work, some preprints~~\citep{song2024slebstreamliningllmsredundancy,kim2024shortened,yang2024laco,men2024shortgpt, gromov2024unreasonable} (SLEB, Shortened Llama, LaCo, ShortGPT and UIDL) were posted, presenting a similar pruning method to ours. However, we have also proposed an additional layer replacement component, which makes our method perform better compared to the aforementioned methods.

\section{\model}

The workflow of the LLM-Streamline framework, shown in Fig.~\ref{fig: workflow} (a), comprises two main steps: layer pruning and layer replacement. First, we prune redundant layers from the LLMs. Then, we train a lightweight network to replace the pruned layers to restore the model's performance.

\subsection{Layer Redundancy in LLMs \label{sec: LLM Redundancy}}

LLMs primarily utilize a Transformer architecture, consisting of a series of Transformer decoder layers. These layers adopt a residual structure, so the effect of each Transformer layer can be viewed as a transformation of the input hidden states. Assuming that the parameters of the $\ell$-th layer $f$ are denoted as $\theta^{(\ell)}$, and its input hidden states are represented by $\boldsymbol{x}^{(\ell)}$, 
%$\boldsymbol{x}^{(\ell)}$ denotes the input hidden states of multiple sequences in a batch, 
the layer $f$ can be expressed as
\begin{align}
\boldsymbol{x}^{(\ell+1)} = \boldsymbol{x}^{(\ell)} + f(\boldsymbol{x}^{(\ell)}, \theta^{(\ell)}).
\label{Eq.layer}
\end{align}
In Eq.~\ref{Eq.layer}, the $\ell$-th layer $f$ contributes a transformation $f(\boldsymbol{x}^{(\ell)}, \theta^{(\ell)})$ to the input $\boldsymbol{x}^{(\ell)}$. Therefore, we assess the importance of each layer in LLMs by evaluating its impact on the input hidden states. We use the cosine similarity $\cos{(\cdot, \cdot)}$ between input $\boldsymbol{x}^{(\ell)}$ and output $\boldsymbol{x}^{(\ell+1)}$ as a metric. Essentially, a higher cosine similarity between the input and output of a layer indicates lower importance, and vice versa. This interpretation arises from the observation that a high cosine similarity suggests the layer's transformation is minimal, making it more amenable to pruning.

To measure the importance of different layers in LLMs, we randomly select samples from the pre-training data (details in Section~\ref{sec: setup}). We then record the hidden states generated by the LLMs for these samples and compute the cosine similarity between the input and output hidden states of each layer. The computation of cosine similarity can be formalized as follows,
\begin{align}
    \cos{(\boldsymbol{x}^{(\ell)}, \boldsymbol{x}^{(\ell + 1)})} = \mathbb{E}_{(\boldsymbol{x}^{(\ell)}_i, \boldsymbol{x}^{(\ell + 1)}_i) \in \mathcal{D}} \left( \frac{1}{L}\sum_{j=1}^{L} \frac{\boldsymbol{x}^{(\ell)}_{i,j} \cdot \boldsymbol{x}^{(\ell + 1)}_{i,j}}{\Vert \boldsymbol{x}^{(\ell)}_{i,j} \Vert \cdot \Vert \boldsymbol{x}^{(\ell + 1)}_{i,j} \Vert} \right),
    \label{Eq.cos}
\end{align}
\noindent where $\mathcal{D}$ denotes the recorded hidden states from different samples, $\boldsymbol{x}^{(\ell)}_i$, $\boldsymbol{x}^{(\ell + 1)}_i \in R^{d \times L}$ denotes the input and output hidden states of the $i$-th sample respectively, $d$ denotes the hidden size and $L$ denotes the sequence length of each sample.

To mitigate the effects of model size and model structure, we conduct experiments on four models OPT-1.3B, OPT-2.7B, OPT-6.7B~\citep{zhang2022opt}, and Llama2-7B~\citep{touvron2023llama}. The results, illustrated in Fig.~\ref{fig:cosine_similarity}, show that there is high cosine similarity between the input and output of several consecutive layers in all models, indicating a low level of importance.

\textbf{Discussion \uppercase\expandafter{\romannumeral1}: Why not use other similarity to measure the importance of layers?} 
%In deep learning, cosine similarity is widely employed to measure vector similarity~\citep{chen2020simple, chen2021exploring, reimers2019sentence}. Alongside it, dot product and Euclidean distance are also utilized, but they additionally consider vector magnitude. Research indicates that Transformer hidden states grow with layers in Pre-Norm architecture\citep{liu2023deja}, biasing dot product towards deeper layers and Euclidean distance towards earlier ones. Thus, we choose magnitude-agnostic cosine similarity.
In deep learning, cosine similarity is widely employed to measure the similarity between two vectors~\citep{chen2020simple, chen2021exploring, reimers2019sentence}. Alongside it, dot product and Euclidean distance are also utilized, but they additionally consider vector magnitude. Current research suggests that the hidden states of Transformers with the Pre-Norm architecture tend to grow as the depth of layers increases~\citep{liu2023deja}. This trend leads to a bias where deeper layers in the model have higher dot product similarity, while earlier layers have smaller Euclidean distances. Consequently, we opt for cosine similarity, which is agnostic to the magnitude of the vectors. 
%We conduct detailed experiments in the Appendix~\ref{sec: similarity metric} to demonstrate that cosine similarity is better similarity metric.
 
\textbf{Discussion \uppercase\expandafter{\romannumeral2}: Why not use perplexity as the metric to measure the importance of layers?} Some concurrent layer pruning work uses perplexity as the metric to measure the importance of layers~\citep{song2024slebstreamliningllmsredundancy, kim2024shortened}. Specifically, they remove each layer one at a time, measuring the change in perplexity of the model on the pre-training data, and eliminate the layer that causes the least change. This process is repeated multiple times to remove several layers. However, we think perplexity is a highly data-sensitive metric, which results in different layers being removed when pruning with different pre-training data. This also results in a situation where, although the perplexity of the pruned model on the pre-training data used for pruning is low, it performs poorly on other datasets. In contrast, the cosine similarity is highly stable and always leads to the same pruned layers on different pre-training data. We conduct detailed experiments in the Appendix~\ref{sec: Comparison} to demonstrate that perplexity is a highly data-sensitive metric and performs poorly on downstream tasks. 

\subsection{Layer Pruning \label{sec: layer pruning}}

As Fig.~\ref{fig:cosine_similarity} shows, the less important layers are often contiguous. Hence, given number of pruned layers $n$ determined by a target sparsity, we remove $n$ contiguous layers by finding the initial layer $\ell^*(n)$ corresponding to the highest cosine similarity for pruning:
%where $n$ is determined by the target sparsity:
\begin{align}
    \ell^*(n) = \underset{\ell}{\arg\max} \, \cos(\boldsymbol{x}^{({\ell})}, \boldsymbol{x}^{({\ell+n})}),
\end{align}
\noindent where we randomly select samples from the pre-training data to compute the cosine similarity between $\boldsymbol{x}^{(\ell)}$ and $\boldsymbol{x}^{(\ell+n)}$, as outlined in Eq.~\ref{Eq.cos}.

\subsection{Layer Replacement \label{sec: layer replacement}}

After the layer pruning process, we aim to replace the pruned layer with a lightweight network that has much fewer parameters. The rationale is that these layers contribute only minor transformations to the input. Therefore, we hypothesize that the cumulative effect of these layers can be approximated by a lightweight network. Specifically, after identifying the initial layer $\ell^*(n)$ for pruning, we use $(\boldsymbol{x}^{(\ell^*)}, \boldsymbol{x}^{(\ell^* + n)})$ as the training data to train the lightweight network using mean squared error (MSE) loss, which can be formalized as follows:
\begin{align}
    \underset{h}{\text{min}} \; \mathbb{E}_{(\boldsymbol{x}^{(\ell^*)}_i, \boldsymbol{x}^{(\ell^* + n)}_i) \in \mathcal{D}} {\rm MSE}(h(\boldsymbol{x}^{(\ell^*)}_i), \boldsymbol{x}^{(\ell^* + n)}_i),
\end{align}
where $h$ denotes the lightweight network, $\mathcal{D}$ denotes the recorded hidden states of samples. 

\textbf{Discussion: Layer Replacement or Fine-Tuning Pruned LLMs?}
Here, we discuss why opt for layer replacement, instead of using common Parameter-Efficient Fine-Tuning (PEFT) methods such as LoRA~~\citep{hu2021lora} and QLoRA~~\citep{dettmers2023qlora} after layer pruning.

First, from the perspective of resource overhead, layer replacement is more adaptable to hardware resource constraints compared to other methods. Fine-tuning the model using the PEFT methods requires storing the model's weights, activation values, and the optimizer state of the PEFT module in the GPU during training. In contrast, layer replacement involves two stages: dataset construction and model training. The first stage only requires storing the model's weight and the forward computation overhead, and the second stage only requires storing of the lightweight network's weight, activation values of lightweight network, and the optimizer state of lightweight network. Therefore, layer replacement can also be implemented under conditions of hardware resource constraints.

Second, layer replacement uses a lightweight network to replace the pruned layer, and distills the knowledge of the pruned layer into the lightweight network using the MSE loss function. Unlike layer replacement, we speculate that training the model after pruning with LoRA is a process of redistributing the function of the pruned layers across the remaining layers. Therefore, substituting the pruned layers with a lightweight network could be a less challenging training task than redistributing the function of the pruned layers across the remaining layers. In the experiments of Section~\ref{sec: layer replacement and lora}, we demonstrate that layer replacement has better performance compared to LoRA.

\begin{table}[t]
\caption{(a) Number of samples in ${\rm TP}, {\rm FN}, {\rm FP}$, and ${\rm TN}$. (b) The PPL standard deviation results ($\times 10^{-3}$) for Llama2-7B and its pruned version on Race-H. \label{tab: quantities and std}}
\begin{minipage}[t]{0.5\textwidth}
\centering
\makeatletter\def\@captype{table}
\renewcommand\arraystretch{1.2}
\resizebox{0.6\linewidth}{!}{
\begin{tabular}{c|cccc}
    \toprule
    Dataset & $\#{\rm TP}$ & $\#{\rm FN}$ & $\#{\rm FP}$ & $\#{\rm TN}$ \\
    \midrule
    C3 & 543 & 257  & 210  & 815 \\
    \midrule
    CHID & 269 & 563  & 177  & 993 \\
    \midrule
    Race-M & 380 & 95 & 129 & 832 \\
    \midrule
    Race-H & 938 & 305 & 353 & 1902  \\
    \bottomrule
\end{tabular}
}
% \caption{Number of samples in ${\rm TP}, {\rm FN}, {\rm FP}$, and ${\rm TN}$.}
\end{minipage}
\begin{minipage}[t]{0.5\textwidth}
\centering
\makeatletter\def\@captype{table}
\renewcommand\arraystretch{1.2}
\resizebox{0.8\linewidth}{!}{
\begin{tabular}{c|cccc}
    \toprule
    Model & ${\rm TP}$ & ${\rm FN}$ & ${\rm FP}$ & ${\rm TN}$ \\
    \midrule
    Llama2-7B & 1.12 & 0.87 & 0.94 & 1.02 \\
    \midrule
    w/ pruning & 1.13 & 0.84 & 0.88 & 0.92  \\
    \bottomrule
\end{tabular}
}
% \caption{The PPL standard deviation results ($\times 10^{-3}$) for the Llama2-7B and its pruned version on Race-H.}
\end{minipage}
\end{table}

\section{Metrics for Evaluating Pruned Models}

Accuracy is the most commonly used metric for evaluating LLMs in classification tasks. However, accuracy may overestimate the performance of the model after compression, since it does not take into account the consistency of the model's answers before and after compression. In this section, we analyze such limitation and propose a novel metric for evaluating compressed models.

\subsection{Shortcoming of Accuracy Metric}

When evaluating the natural language understanding capabilities of LLMs, most existing benchmarks frame the task as a classification task. 
A classification task with $k$ choices and comprising $N$ samples is denoted as $\mathcal{T} = \{\left(x_i, c_{i,1}, c_{i,2}, ..., c_{i,k}, y_i \right)\}_{i=1}^{N}$, where $x_i$ represents the question in the $i$-th sample, $c_{i,j}$ represents the $j$-th choices, and $y_i$ represents the correct choice.
The input to the classification task consists of a question accompanied by multiple choices, and the LLM is required to select the correct answer from these choices. Typically, each choice is concatenated with the question to form multiple sentences, and the perplexity (PPL) of each sentence is computed. The choice corresponding to the sentence with the lowest PPL is selected as the answer.

Typically, model pruning results in decreased model performance. However, when we evaluate the model pruned by the method described in Sec~\ref{sec: layer pruning}, we unexpectedly observe the accuracy of the pruned model has been improved on some classification tasks.  
We define $\mathcal{M}$ to denote the original LLM, $\bar{\mathcal{M}}$ to denote the compressed LLM, and $\hat{y}(\mathcal{M})$ to denote the choice predicted by the model $\mathcal{M}$.
To further investigate this phenomenon, we analyze the experimental results using the confusion matrix~~\citep{li2024evaluating}. Specifically, we count the number of samples and average standard deviation (std) for the PPL of the samples for each term of the confusion matrix. The calculation of the std for the PPL of the $i$-th sample is defined as follows:
\begin{align}
\small
 \mathrm{PPL}_{i,j} = \mathrm{PPL}(\mathcal{M}(x_i, c_{i,j})), {\mathrm{PPL}}_i = \frac{\sum_{j=1}^{k}\mathrm{PPL}_{i,j}}{k}, \mathrm{std}_i = \sqrt{\frac{\sum_{j=1}^{k}(\mathrm{PPL}_{i,j} - {\mathrm{PPL}}_i)^2}{k-1}},
\end{align}
\noindent where ${\rm PPL}_{i,j}$ denotes the PPL for the sentence created by question $x_i$ and choice $c_{i,j}$ of the model before pruning, ${\rm std}_i$ denotes the std for PPL of the $i$-th sample. A higher $\mathrm{std}_i$ value indicates the LLM exhibits greater confidence in answering the question $x_i$.

Each term of the confusion matrix is defined as follows,

\begin{itemize}
    \item ${\rm TP} \left[\hat{y}_i({\mathcal{M}}) = y_i \wedge  \hat{y}_i(\bar{\mathcal{M}}) = y_i\right]$ is a set of samples where the model answers correctly both before and after pruning.

    \item ${\rm FN}  \left[\hat{y}_i({\mathcal{M}}) = y_i \wedge \hat{y}_i(\bar{\mathcal{M}}) \neq y_i \right]$ is a set of samples where the model answers correctly before pruning but incorrectly after pruning.

    \item ${\rm FP} \left[ \hat{y}_i({\mathcal{M}}) \neq y_i  \wedge  \hat{y}_i(\bar{\mathcal{M}}) = y_i \right]$ is a set of samples where the model answers incorrectly before pruning but correctly after pruning.

    \item ${\rm TN} \left[\hat{y}_i({\mathcal{M}}) \neq y_i \wedge \hat{y}_i(\bar{\mathcal{M}}) \neq y_i \right]$ is a set of samples where the model answers incorrectly both before and after pruning.

\end{itemize}

Table~\ref{tab: quantities and std} presents the counts of samples in TP, FN, FP, TN in sevaral datasets, and also the PPL standard deviation in Race-H dataset.
We can observe that the std for ${\rm TP}$ and ${\rm TN}$ is significantly higher than that for ${\rm FN}$ and ${\rm FP}$. 
This indicates that the model is more uncertain about the ${\rm FN}$ and ${\rm FP}$ samples. In addition, the samples in ${\rm FP}$ constitute a considerable proportion of the total samples, implying that the model may guess the correct answer for a significant portion after pruning. This phenomenon suggests that the accuracy metric may overestimate the performance of the compressed model.
%We can observe that samples in ${\rm FP}$ constitute a considerable proportion of the total samples, and the std for ${\rm TP}$ and ${\rm TN}$ is significantly higher than that for ${\rm FN}$ and ${\rm FP}$. This indicates that the model is more uncertain about the ${\rm FN}$ and ${\rm FP}$ samples, implying that the model may guess the correct answer for a significant portion of these samples after pruning. This phenomenon suggests that the accuracy metric may overestimate the performance of the compressed model.

\subsection{Stability Metric}

We propose a novel metric stability to evaluate the performance of LLMs after pruning, i.e.,
\begin{gather}
    \mathrm{Stability}(\mathcal{M}, \bar{\mathcal{M}}) = \frac{\sum_{i=1}^{N}\left( \exp{(\mathrm{std}_i)} \cdot \mathbbm{1}_{\left[ i \in {\rm TP} \cup {\rm TN} \right]} \right)}{\sum_{i=1}^{N}\exp{(\mathrm{std}_j})},
\end{gather}

where the identifier $\mathbbm{1}_{\left[ i \in {\rm TP} \cup {\rm TN} \right]}$ is used to indicate whether the $i$-th sample belongs to ${\rm TP}$ and ${\rm TN}$. We use ${\rm std}_i$ as the weight of the  $i$-th sample. Because the ${\rm std}$ of different samples varies significantly, to mitigate the influence of samples with excessively large standard deviations, we apply the $\exp$ function to moderate the weight differences among samples. Different from accuracy, stability focuses on the model’s confidence in its answers and the consistency between the model before and after pruning on a task, aligning more closely with the goal of model pruning, i.e., ensuring the pruned model remains as similar as possible to the original model.

\section{Experiments}

In this section, we first compare our proposed method, LLM-Streamline, with several popular pruning methods to demonstrate its effectiveness (\ref{sec: main results}). Next, we analyze the impact of different sizes and structures of lightweight networks on model performance (\ref{sec: lightweight network}) and evaluate performance under various pruning ratios (\ref{sec: pruning ratio}). Finally, we compare our layer replacement approach with the well-known PEFT method, LoRA~\citep{hu2021lora} (\ref{sec: layer replacement and lora}), showing that layer replacement offers superior performance and reduced memory overhead.

\subsection{Setup \label{sec: setup}}

We conduct experiments on popular open-source LLMs, including Llama2-7B and Llama2-13B~\citep{touvron2023llama}. Following previous work~\citep{men2024shortgpt, yang2024laco}, we perform experiments with a 25\% pruning rate and extract data from pre-training dataset SlimPajama which contains data from different domains for layer pruning and layer replacement. Sheared LLaMa~~\citep{xia2023sheared} finds that the performance degradation of pruned models varies across different domains, and proposes determining the distribution of data from different domains based on the degree of performance degradation. Therefore, we randomly sample the data based on the distribution used by Sheared LLaMa~~\citep{xia2023sheared}, finally constructing the dataset containing 30,000 pieces of data. We randomly select 500 samples from this dataset and input them into LLMs, generating Fig.~\ref{fig:cosine_similarity}, and use these 500 data samples for layer pruning. All 30,000 pieces of data are used to train the lightweight network. 
We utilize two types of lightweight networks: a Feed-Forward Neural Network (FFN), referred to as Ours (FFN), and a Transformer Layer, referred to as Ours (Layer). The FFN is randomly initialized, while the Transformer Layer inherits the parameters from the first pruned layer. Additionally, we explore a purely pruning approach without incorporating any lightweight network, denoted as Ours (None).
Further experimental details are available in the Appendix~\ref{sec: lightweight network training details}.

\subsection{Benchmark \label{sec: benchmark}}

%To comprehensively assess the changes in LLM capabilities before and after pruning, 
%we follow the evaluation method of LaCo~~\citep{yang2024laco}. 
We use 12 natural language understanding benchmarks for evaluation: \textbf{CMNLI}~~\citep{xu2020clue}, \textbf{HellaSwag}(HeSw)~~\citep{zellers2019hellaswag}, \textbf{PIQA}~~\citep{bisk2020piqa},\textbf{CHID}~~\citep{zheng2019chid}, \textbf{WSC}~~\citep{levesque2012winograd},\textbf{CommonSenseQA}(CoQA)~~\citep{talmor2018commonsenseqa}, \textbf{BoolQ}~~\citep{clark2019boolq},\textbf{MMLU}~~\citep{hendrycks2020mmlu}, \textbf{CMMLU}~~\citep{li2023cmmlu},\textbf{Race-High/Middle}~~\citep{lai2017race}, \textbf{C3}~~\citep{sun2020C3}. The tasks in these benchmarks are formalized as classification tasks, so we refer to these benchmarks as classification benchmarks. For these benchmarks, we use both accuracy and stability as metrics for evaluating the models. Additionally, we include 3 benchmarks: \textbf{XSum}~~\citep{narayan2018xsum},  \textbf{GSM8K}~~\citep{cobbe2021gsm8k} and \textbf{StrategyQA}~~\citep{geva2021did}, to demonstrate the LLM’s performance on generation tasks after pruning. 
We refer to these tasks as generation benchmarks. Following the evaluation framework of  OpenCompass~\citep{contributors2023opencompass}, we use accuracy to evaluate \textbf{StrategyQA} and \textbf{GSM8K}, and use ROUGE1 to evaluate \textbf{Xsum}.

\subsection{Baseline \label{sec: baseline}}

We compare several pruning methods that prune the attention heads, the filters of the FFN layer, and the hidden dimension, as well as the concurrent layer-pruning methods LaCo. In addition, \textbf{ShortGPT} and \textbf{UIDL}~\citep{men2024shortgpt, gromov2024unreasonable} can be considered as a variant of our approach, i.e., Ours (None). We also discuss layer pruning methods which use perplexity as the metric, such as \textbf{SLEB}~\citep{song2024slebstreamliningllmsredundancy}, in Appendix~\ref{sec: Comparison}.

\textbf{LLM-Pruner}~\citep{ma2023llmpruner} prunes attention heads, FFN layer filters, and hidden dimensions by using gradients and activations to estimate the importance of these modules.

\textbf{SliceGPT}~\citep{ashkboos2024slicegpt} prunes hidden dimensions. 
%It inserts dimensionality reduction matrices into the model and employs Principal Component Analysis (PCA) to initialize and eliminate unimportant components within these matrices. The reduced-dimensional matrices are subsequently merged with the model's original weight matrices, thereby reducing the model’s size.
It inserts dimensionality reduction matrices into the model and employs Principal Component Analysis (PCA) to initialize and compress the matrices, and then merge them with the original weight matrix to reduce the model’s size.

\textbf{LaCo}~\citep{yang2024laco} prunes layers by dividing the layers into groups, each consisting of multiple consecutive layers, and compresses them separately, whereas our method simply compresses a piece of consecutive layers. LaCo merges consecutive layers by averaging their parameters whereas we train an additional lightweight network to replace these layers.

\begin{table}[t]
\caption{Accuracy of pruning methods on classification benchmarks. ``*'' indicates that we refer to the results in the original paper. Retained performance (RP) represents the percentage of the original model’s performance retained by the pruning method.}
	\small
	\centering
\renewcommand\arraystretch{0.95}
\resizebox{\linewidth}{!}{

	\begin{tabular}{c|c|c|cccccccccccc|cc}
		\toprule
		\multirow{2}{*}{LLM} & \multirow{2}{*}{Method}  & \multirow{2}{*}{Ratio}& \multicolumn{12}{c|}{Benchmarks}& \multirow{2}{*}{Average} & \multirow{2}{*}{RP} \\  
		&   & & C3 & CMNLI&CHID&BoolQ&WSC&CoQA&HeSW&PIQA&Race-M&Race-H &MMLU & CMMLU& &\\
		\midrule
		\multirow{7}{*}{Llama2-7B} 
  & Dense  & 0.00\% &43.8 &33.0 &41.6  &70.8 &37.5  &66.7 &71.3 &78.1  &33.1  &35.5&46.8& 31.8  &49.2 &100.0
		\\
		& LLMPruner & 24.8\% &29.7  &\underline{33.4}&\underline{28.4}  &58.7 &\underline{40.4}  &48.5 &54.6 &\textbf{72.0} &22.9 &22.0&25.3   &25.0    &38.4 &78.0\\
		& SliceGPT  & 25.4\%& 31.5 &31.6 &18.5 &59.9 &\textbf{43.3} & 49.6& 47.5&  68.3 &27.0& 29.4& 28.8 & 24.8  &38.4 &78.0   \\
		& LaCo*  & 27.0\%&39.7  &\textbf{34.4}&\textbf{36.1}  &64.1 &\underline{40.4} &45.7 &55.7 &69.8&23.6   &22.6&26.5  &25.2  &40.3 &81.9\\
        % & ShortGPT  & 27.1\%&39.6  &33.0&24.7  &74.7 &52.5 &48.0 &53.0 &66.4&35.2   &32.3&44.0  &32.3  &44.6 &90.7\\
        % & ShortGPT*  & 27.1\%&41.3  &33.1&22.3  &65.2 &36.5 &48.8 &58.2 &66.9&35.5   &36.9&45.4  &29.1  &43.3 &88.0\\
        & Ours (None)  & 24.0\%&40.2  &\textbf{34.4}&21.5  &\underline{67.3} &\underline{40.4} &51.7 &59.7 &69.0&\underline{35.2}   &34.7&44.6  &28.9  &44.0 &89.4\\
        & Ours (FFN)  & 25.0\%&\underline{40.7}  &33.0& 22.8 &65.9 &38.5 &\textbf{60.6} &\textbf{61.2} &71.2&\textbf{38.0}   &\textbf{38.7}&\textbf{47.0}  &\textbf{31.7}  &\textbf{45.8} &\textbf{93.1}\\
        & Ours (Layer) & 24.0\%&\textbf{43.3}  &33.0&24.1 &\textbf{67.5} &36.5 &\underline{59.2} &\underline{61.1} &\underline{71.5}&34.8   &\underline{37.0}&\underline{45.5}  &\underline{29.4}  &\underline{45.2} &\underline{91.9}\\
            
            \midrule
            \multirow{7}{*}{Llama2-13B} 
  & Dense  & 0.00\% &47.5 & 33.0 & 47.2 & 71.5 & 51.0 & 66.8 & 74.8 & 79.8 & 60.0 & 58.1 & 55.8 & 38.7 & 57.0 & 100.0\\
             & LLMPruner &24.4\% &29.5&\textbf{33.0}&29.5&58.0&47.1&43.7&54.7&72.7&21.9&22.5&25.2&24.9&38.6&67.7 \\
             & SliceGPT &23.6\% &38.6&30.5&18.3&37.8&42.3&38.3&45.6&61.9&24.0&25.0&30.6&25.6&34.9&61.2 \\
             & LaCo* & 24.6\%
&44.9 & \underline{32.9} & \textbf{40.1} & 64.0 & \underline{52.9} & 52.7 & 64.4 & \underline{74.3} & 56.6 & 54.5 & 45.9 & 32.6  &51.3  &90.0  \\
             & Ours (None)  & 24.6\%&\textbf{47.0} & \textbf{33.0} & 36.5 & 62.3 & \textbf{64.4} & 58.8 & 66.6 & 73.5 & \textbf{60.2} & \textbf{58.3} & 54.8 & 38.4 & \textbf{54.5} & \textbf{95.6}\\
             & Ours (FFN) & 25.4\%&\underline{45.8} & \textbf{33.0} & 37.1 & \textbf{67.4} & 37.5 & \textbf{64.4} & \underline{67.9} & 74.0 & \underline{58.6}& \underline{58.2} & \textbf{55.7} & \underline{38.6} & \underline{53.2} & \underline{93.3}\\
             & Ours (Layer) & 24.6\%& 45.7 & \textbf{33.0} & \underline{38.0} & \underline{66.2} & 36.5 & \underline{63.8} & \textbf{69.1} & \textbf{75.1} & 58.0 & 57.4 & \underline{55.1} & \textbf{39.2} & 53.1 & 93.2\\          
		\bottomrule
	\end{tabular}
 }
	\label{tab:llm_comparison_all_7B_PPL_A}
\end{table}

\begin{table}[t]
\caption{Stability of pruning methods on classification benchmarks. The stability of the original model is 1.0, because stability is measured by comparing the prediction results of the original model.}
	\small
	\centering
\renewcommand\arraystretch{0.95}
\resizebox{\linewidth}{!}{

	\begin{tabular}{c|c|c|cccccccccccc|c}
		\toprule
		\multirow{2}{*}{LLM} & \multirow{2}{*}{Method}  & \multirow{2}{*}{Ratio}& \multicolumn{12}{c|}{Benchmarks}& \multirow{2}{*}{Average} \\  
		&   & & C3 & CMNLI&CHID&BoolQ&WSC&CoQA&HeSW&PIQA&Race-M&Race-H &MMLU & CMMLU& \\
		\midrule
		\multirow{5}{*}{Llama2-7B} 
    		& LLMPruner & 24.8\% &72.8  &\underline{94.0}&\textbf{74.1}  &70.8 &87.5  &71.0 &79.9 &\textbf{86.8} &52.4 &55.2&53.3   &65.9    &72.0 \\
    		& SliceGPT & 25.4\% &53.2  &35.4&53.3  &77.1 &80.8  &75.3 &71.6 &78.7 &\textbf{90.7} &\textbf{85.3}&60.3 &56.7    &68.2 \\
            % & ShortGPT* & 27.1\% &74.5  &99.7&63.0  &81.9 &95.2  &66.1 &76.0 &77.9 &84.4 &81.2&84.0   &69.0    &79.4 \\
            & Ours (None) & 24.0\% &\underline{76.6}  &38.7&\underline{65.3}  &81.4 &87.5  &74.7 &80.7 &81.0 &73.7 &67.9&80.1   &70.8    &73.2\\
            & Ours (FFN) & 25.0\% &\textbf{79.8}  &\textbf{100}&64.1  &\underline{83.1} &\underline{93.3}  &\underline{80.7} &\underline{84.7} &84.6 &\underline{85.1} &\underline{79.0}& \textbf{87.5}  &\textbf{82.5 }   &\textbf{83.7} \\
            & Ours (Layer) & 24.0\% &\textbf{79.8}  &\textbf{100}& 64.4 &\textbf{86.3} &\textbf{95.2}  &\textbf{81.7} &\textbf{85.3} &\underline{85.6} &81.8 &\underline{79.0}&\underline{82.4}   &\underline{71.0}    &\underline{82.7} \\
            
            \midrule
            \multirow{5}{*}{Llama2-13B} 
            & LLMPruner & 24.4\% 
            &71.6&\textbf{100}&69.2&70.5&\underline{65.4}&69.5&77.8&86.7&42.3&35.6&48.1&52.3&65.8\\
            & SliceGPT & 23.6\% 
            &62.2&39.5&51.4&27.1&\textbf{68.3}&65.5&64.9&75.6&45.3&43.4&52.7&52.9&54.1\\
            & Ours (None) & 24.6\% &84.2 & \underline{99.9} & 71.8 & 77.4 & 46.2 & 82.2 & 85.7 & 86.5 & 83.3 & \textbf{83.6} & 89.1 & 83.8 & 81.1 \\
            & Ours (FFN) & 25.4\% &\underline{85.7} & \textbf{100} & \underline{72.5} & \underline{79.8} & 59.6 & \textbf{89.2} & \underline{89.4} & \underline{89.7} & \textbf{84.8} & \underline{83.3} & \textbf{93.6} & \textbf{90.7} & \textbf{84.9}\\
            & Ours (Layer) & 24.6\% &\textbf{87.4} & \textbf{100} & \textbf{74.1} & \textbf{81.3} & 58.6 & \underline{89.0} & \textbf{90.5} & \textbf{90.5} & \underline{84.2} & 83.0 & \underline{92.5} & \underline{85.5} & \underline{84.7} \\
		\bottomrule
	\end{tabular}
 }
	\label{tab:llm_comparison_all_7B_PPL_S}
\end{table}

\begin{table}[t]
\caption{Evaluations on generation benchmarks. ``*'' indicates that we refer to the results in the original paper.}
\small
\centering
\renewcommand\arraystretch{0.8}
\resizebox{0.75\linewidth}{!}{
\begin{tabular}{c|c|c|ccc|cc}
    \toprule

    \multirow{2}{*}{LLM} & \multirow{2}{*}{Method}  & \multirow{2}{*}{Ratio}& \multicolumn{3}{c|}{Benchmarks}& \multirow{2}{*}{Average} & \multirow{2}{*}{RP} \\  
    &   & & Xsum & GSM8K & StrategyQA& &\\
    \midrule
    \multirow{7}{*}{Llama2-7B} 
& Dense  & 0.00\% &19.4 &16.5 &60.2 &32.0 &100.0
    \\
    & LLMPruner & 24.8\% &16.4 &0.61 &44.2 &20.4 &63.8\\
    & SliceGPT  & 25.4\%&12.4 &\textbf{3.34} &45.7 &20.5 &64.1   \\
    & LaCo*  & 27.1\%&15.6 &- &- &- &-\\
    % & ShortGPT*  & 27.1\%&0.7 &1.70 &35.1 &12.5 &39.1 \\
    & Ours (None)  & 24.0\%&14.8 &1.97 &41.8 &19.5 &60.9 \\
    & Ours (FFN) & 25.0\% &\underline{18.6} &\underline{2.16} &\underline{46.5} &\underline{22.4} &\underline{70.0}\\
    & Ours (Layer) & 24.0\% &\textbf{20.2} &1.82 &\textbf{52.1} &\textbf{24.7} &\textbf{77.2}\\
    \midrule
    \multirow{7}{*}{Llama2-13B} 
    & Dense  & 0.00\% &23.7 &29.0 &58.1 &36.9 &100.0
    \\
    & LLMPruner & 24.4\% &17.5&1.9&43.7&21.0&56.9\\
    & SliceGPT & 23.6\% &5.0&1.9&38.3&15.1&40.9\\
    & LaCo* & 24.6\% &14.5&-&-&-&-\\
    & Ours (None) & 24.6\% &17.7 &2.35 &46.0 &22.0 &59.6\\
    & Ours (FFN) & 25.4\%  &\underline{21.4} &\underline{4.10} &\textbf{59.6} &\textbf{28.4} &\textbf{77.0}\\
    & Ours (Layer) & 24.6\%  &\textbf{21.8} &\textbf{4.70} &\underline{57.3} &\underline{27.9} &\underline{75.6}\\

    \bottomrule
\end{tabular}
}
\label{tab:llm_comparison_all_7B_GEN}
\end{table}

\begin{table}[t]
\caption{Comparison of different lightweight networks on classification benchmarks in terms of average accuracy and stability metrics, where ``$\dag$'' indicates that the intermediate size of the added lightweight network is half that of the default LLM's intermediate size.}
	\small
	\centering
\renewcommand\arraystretch{0.8}
\resizebox{\linewidth}{!}{

	\begin{tabular}{c|cccccccc}
    \toprule
    & Layer-Random & Layer-First & Layer-Last & Layer-Avg & FFN$^\dag$ & FFN & SwiGLU$^\dag$ & SwiGLU \\
    \midrule
    Accuracy & 45.1 & 45.2 & 45.6 & 44.4 & \textbf{46.0} & \underline{45.8} & 43.8 & 44.2 \\
    \midrule
    Stability & 81.2 & 82.7 & 81.9 & 79.2 & 80.7 & \textbf{83.7} & 82.6 & \underline{83.3} \\
    \bottomrule
\end{tabular}
 }
	\label{tab:different_lightweightmodel_PPL}
\end{table}

% \begin{table}[t]
% \small
% \centering
% \renewcommand\arraystretch{1.2}
% \resizebox{\linewidth}{!}{
% \begin{tabular}{c|ccccc}
%     \toprule
%     LLM & Metric & Pruned layers & Perplextity on SlimPajama &Average Performance on Classification Benchmarks & Average Performance on Generation Benchmarks \\
%     \midrule
%     \multirow{3}{*}{Llama2-7B} 
%     &- &- &6.23 &49.2 &32.0\\
%     &Cosine Similarity &22,23,24,25,26,27,28,29 &19.7 &\textbf{44.0} &\textbf{19.5}\\
%     &Perplexity &6,7,8,9,10,11,12,13 &\textbf{13.3} &32.3 &7.95\\
%     \bottomrule
        
% \end{tabular}
% }
% \caption{Comparison of perplexity and cosine similarity.}
% 	\label{tab:ppl_cosine}
% \end{table}

\subsection{Main Results \label{sec: main results}}

We present accuracy and stability for different methods on the classification benchmarks in Table~\ref{tab:llm_comparison_all_7B_PPL_A} and Table~\ref{tab:llm_comparison_all_7B_PPL_S}, respectively, and Table~\ref{tab:llm_comparison_all_7B_GEN} for the generative benchmarks. The results demonstrate that our proposed LLM-Streamline consistently outperforms the baseline methods. Specifically, in classification tasks, LLM-Streamline surpasses LLM-Pruner by 7\% in accuracy and 12\% in stability on Llama2-7B, and by 16\% in accuracy and 19\% in stability on Llama2-13B. LLM-Streamline also surpasses LaCo by 5\% in accuracy on Llama2-7B. 
For generation tasks, LLM-Streamline retains nearly 77\% of Llama2-7B and Llama2-13B's capabilities, significantly outperforming other pruning methods. We find that almost all of the pruning methods fail on the GSM8K dataset. However, sufficient training can gradually restore the model’s performance on math tasks, and the specific experimental results are shown in Table~\ref{tab: training data volumes} of  Appendix~\ref{Sufficient Post-training}.

Additionally, comparing the average stability (Average) in Table~\ref{tab:llm_comparison_all_7B_PPL_S} with the retrained performance (RP) in Table~\ref{tab:llm_comparison_all_7B_PPL_A} reveals that stability is often much lower than accuracy. We also observe that accuracy on Race-M and Race-H even increases after model pruning. Furthermore, we find that without using any lightweight network, Llama2-13B achieves the highest accuracy on classification benchmarks, but its stability on classification benchmarks and performance on generation benchmarks are lower. These results indicate that pruned models tend to make correct guesses on some classification questions that they are uncertain, highlighting the limitations of accuracy as a sole measure of pruning method performance. 
We also conduct experiments on OPT-1.3B, OPT-2.7B, OPT-6.7B, Baichuan-7B, Baichuan-13B, Baichuan2-7B, Baichuan2-13B~\citep{yang2023baichuan}, Llama3.1-8B, Llama3.1-70B~\citep{dubey2024llama} and Mixtral-8x7B-v0.1~\citep{jiang2024mixtral}. 
Details can be found in Appendix~\ref{sec: Result of OPT-6.7B and OPT-13B}, Appendix~\ref{sec: Result of Baichuan-7B and Baichuan-13B}, Appendix~\ref{sec: Result of Baichuan2-7B and Baichuan2-13B}, Appendix~\ref{sec: Llama3.1}, Appendix~\ref{sec: MOE} 
and Appendix~\ref{sec: Result of OPT-1.3B and OPT-2.7B}. In addition, we compare the performance of LLM-Streamline with other methods at a higher pruning ratio of approximately 50\%, and the results can be found in Appendix~\ref{sec:0.5}. 
%We also evaluate the inference speed of the models pruned using each method, and the results can be found in  Appendix~\ref{sec: speed time}. 

% \subsection{Comparison of Different Metrics}
% To investigate which is more reasonable between cosine similarity and other metrics for measuring the impact of each layer on the model’s overall performance, such as the change in perplexity, we randomly select 500 samples from SlimPajama-6B and input them into the model to calculate the perplexity. Subsequently, we sequentially remove layers from the model, recalculate the perplexity after each removal, and used the resulting changes as a basis for assessing the importance of individual layers. We directly remove some layers from Llama2-7B based on perplexity and cosine similarity, respectively. As shown in Table \ref{tab:ppl_cosine}, using perplexity as the pruning metric significantly reduces the perplexity on SlimPajama-6B, but performs poorly on some downstream tasks. Detailed results can be found in Appendix~\ref{sec: Comparison}.
%We find that this is due to a gap between the pre-training data and the downstream task data. When perplexity is used as the pruning metric, it can lead to a significant increase in the model's perplexity on some downstream tasks. We discuss this in more detail in Appendix~\ref{sec: Comparison}.

\subsection{Impact of Different Lightweight Networks \label{sec: lightweight network}}

\begin{figure}[t]
\centering
\begin{minipage}[t]{0.4\textwidth}
\centering
    \includegraphics[width=\textwidth]{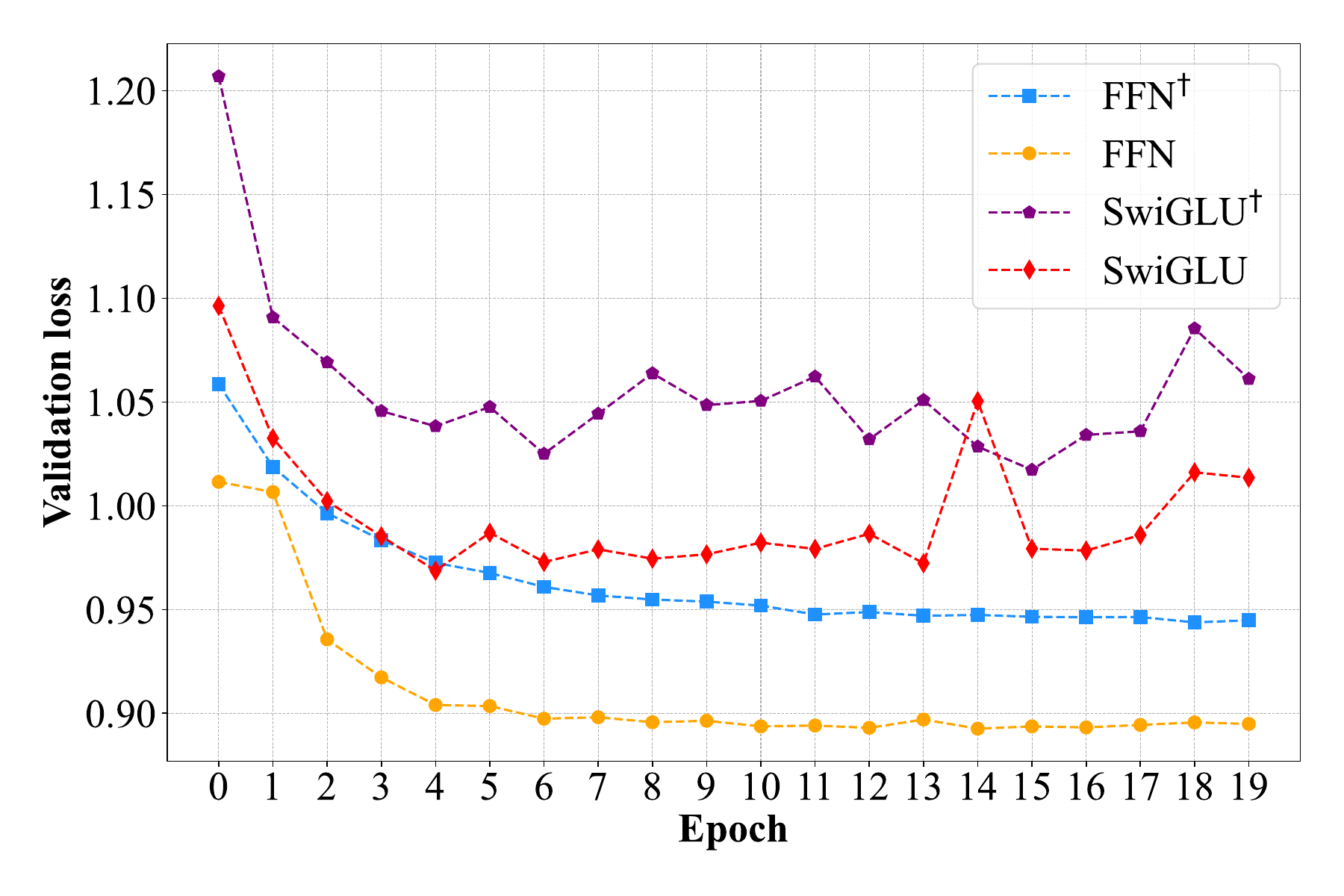}
    % \caption{Valid loss curves of FFN structure and SwiGLU structure}
\end{minipage}
%\hfill
\begin{minipage}[t]{0.4\textwidth}
\centering
    \includegraphics[width=\textwidth]{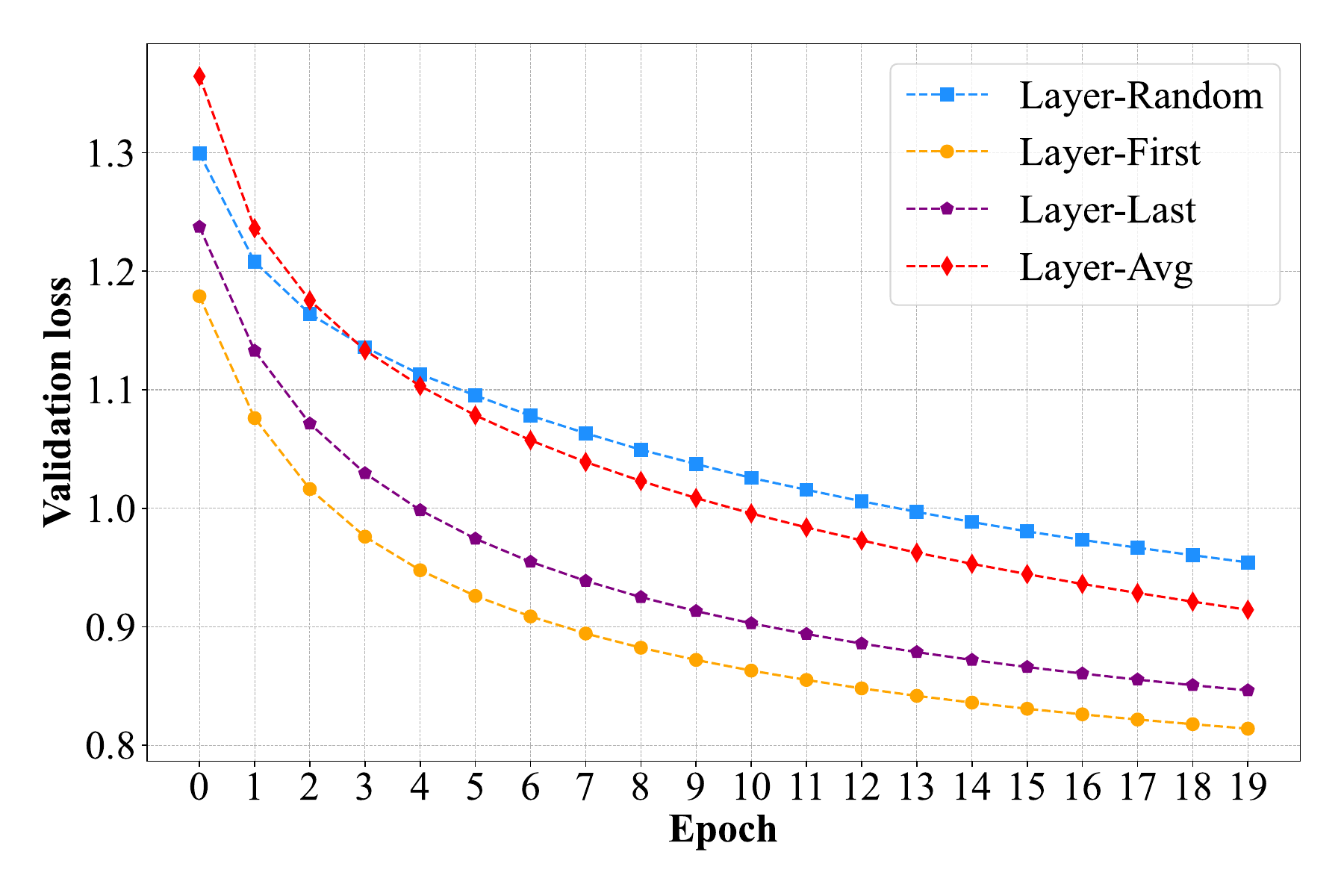}
%     \caption{Valid loss curves of Transformer Layer structure}
% \label{fig: valid loss layer}
\end{minipage}
\caption{Validation loss curves during training of (a) FFN and SwiGLU; (b) Transformer layer. \label{fig: valid loss}}

\end{figure}

\textbf{While FFN achieves the best result, Transformer layer still has performance potential.} 
%To investigate the impact of the model size, architecture, and initialization methods of lightweight networks on the pruned model’s performance, 
We perform experiments with Llama2-7B using various lightweight network, including Feed-Forward Neural Networks (FFN), SwiGLU-based Feed-Forward Neural Networks (SwiGLU), and Transformer layers.
We also explore various initialization methods for the Transformer Layer, including random initialization (Layer-Random), inheritance of the first pruned layer (Layer-First), inheritance of the last pruned layer (Layer-Last), and averaging the pruned layers (Layer-Avg). The average accuracy and stability metrics across all the classification benchmarks are presented in Table~\ref{tab:different_lightweightmodel_PPL}, with detailed results on each benchmark available in Appendix~\ref{Detailed result of different lightweight models}. The results show that FFN achieves the best results. 
Meanwhile, for the Transformer Layer, inheriting the pruned first layer yields the best results. In contrast, the performance of Layer-Avg, inspired by LaCo, shows that averaging weights does not achieve the same effectiveness as the pruned first layer. In addition, we plot the validation loss curves for different lightweight networks, as shown in Fig.~\ref{fig: valid loss}. We can observe that FFN and SwiGLU have already converged by the 10th epoch, whereas the loss of Transformer Layer is still decreasing. This indicates that the Transformer layer still has potential, and further training could yield better results, but this would require more computing resources.
% We also tried different lightweight networks, including Feed-Forward Neural Network (FFN), SwiGLU-based Feed-Forward Neural Network (SwiGLU), and Transformer Layer (Layer).

\subsection{Impact of Different Pruning Ratios \label{sec: pruning ratio}}

\textbf{The performance of the pruned model is linearly correlated with the number of parameters at modest pruning ratios.} 
To verify the model's performance at various modest pruning ratios, we evaluate our method not only at the approximately 25\% pruning ratio but also at ratios of around 2\%, 11\%, and 20\% on Llama2-7B.
The average stability and accuracy metrics across all the classification benchmarks are shown in Fig.~\ref{fig: stability and accuracy at different pruning ratios}, with details on each benchmark presented in Appendix~\ref{Detailed result of pruning ratio}. By comparing the performance of the original Llama2-7B, TinyLlama-1.1B, OpenLlama-3B-v2, and Llama2-7B pruned at various ratios, we observe a linear correlation between the performance of both the pruned models and the pre-trained original models relative to the number of parameters. This suggests that the performance of models pruned using our method is comparable to that of pre-trained models with the same number of parameters.

\subsection{Comparison of Layer Replacement and LoRA \label{sec: layer replacement and lora}}

\textbf{Layer Replacement outperforms LoRA in both performance and GPU memory consumption.} 
We compare the performance of layer replacement with LoRA. Since layer replacement is trained based on hidden states with a different training objective than LoRA, we additionally train one epoch using the language model loss for layer replacement when comparing it with LoRA. The training details can be found in the Appendix~\ref{sec: post-training details}. 
For layer replacement, we freeze the original model's weights and train only the lightweight network. In the case of LoRA, we set the rank to 128 to align the number of parameters trained with those of the lightweight networks. We randomly extract 30,000 samples from SlimPajama-6B for layer replacement training and also test with the entire dataset to evaluate the limited impact of extensive data on performance (details in Appendix~\ref{Sufficient Post-training}). For LoRA, we use 300,000 samples from SlimPajama-6B.
Table~\ref{tab: PostTraing_PPL} presents the average accuracy and stability across all classification benchmarks, with detailed results available in Appendix~\ref{detail result of continual fine-tuning}. 
The findings indicate that layer replacement surpasses LoRA in both accuracy and stability, while also requiring significantly less GPU memory and training data.

\begin{figure}[t]
\centering
\begin{minipage}{0.4\textwidth}
\centering
\includegraphics[width=\textwidth]{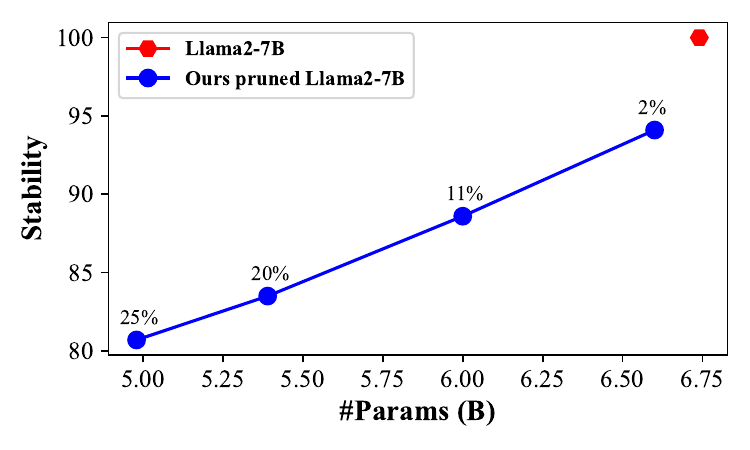} % 插入图片，设置宽度为文本宽度的50%
% \caption{Stability of the pruned Llama-7B at different pruning ratios.} % 添加标题
\end{minipage}
\begin{minipage}{0.4\textwidth}
\centering
\includegraphics[width=\textwidth]{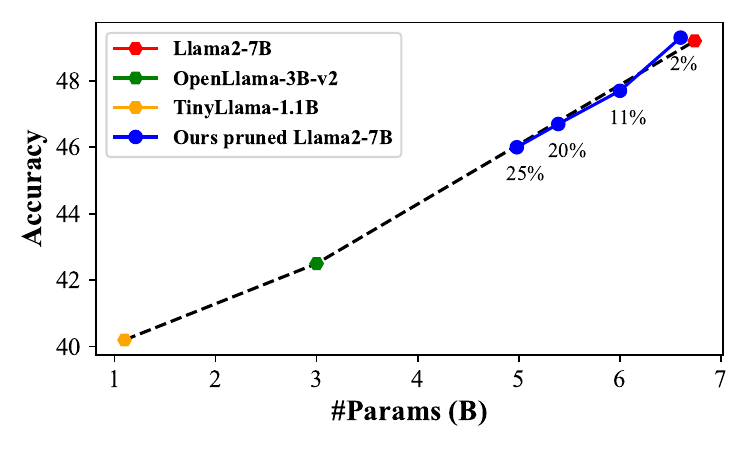} % 插入图片，设置宽度为文本宽度的50%
% \caption{Accuracy of the pruned Llama-7B at different pruning ratios.} % 添加标题
% \label{fig: accuracy at different pruning ratios} % 定义标签
\end{minipage}
\caption{(a) Stability of the pruned Llama2-7B at different pruning ratios.
%of approximately 2\%, 11\%, 20\%, and 25\%. 
%Stability is calculated only for the compressed model compared to the original, so it cannot be computed for TinyLlama-1.1B and OpenLlama-3B-v2. 
(b) Accuracy of the pruned Llama2-7B at different pruning ratios, compared to the original Llama2-7B, OpenLlama-3B-v2, and TinyLlama-1.1B. Metrics are averaged across classification benchmarks.\label{fig: stability and accuracy at different pruning ratios} }
\end{figure}

\begin{table}[t]
\caption{Comparison of layer replacement and LoRA on classification benchmarks in terms of average accuracy metrics across all benchmarks, where ``$\dag$'' indicates that the intermediate size of the added lightweight network is half that of the default LLM's intermediate size.}
	\centering
\renewcommand\arraystretch{0.5}
\resizebox{\linewidth}{!}{

	\begin{tabular}{c|ccccccc|c}
    \toprule
    & Layer-First & Layer-Last & Layer-Avg & FFN$^\dag$ & FFN & SwiGLU$^\dag$ & SwiGLU & LoRA \\ % & LoRA*
    \midrule
    Accuracy & \underline{46.7} & \textbf{46.8} & 46.2 & 45.8 & 46.3 & 44.4 & 45.5 & 44.5 \\ % 44.6
    \midrule
    Stability & \textbf{85.7} & \underline{85.6} & 83.9 & 83.4 & 85.2 & 84.7 & 84.7 & 82.1 \\ % &82.6
    \midrule
    GPU Memory (G) & 27.8 &27.8 &27.8 & \underline{25.6} &27.0 &\textbf{25.3} &26.4 &56.4 \\
    \bottomrule
\end{tabular}
 }
	\label{tab: PostTraing_PPL}
\end{table}

\begin{table}[H]
\caption{Comparison of concurrent layer pruning methods, with the metric indicating the importance of layers. Shortened Llama consists of two training stages: initial continual pre-training on the SlimPajama dataset, followed by LoRA fine-tuning on the Alpaca dataset.
 }
	\centering
\renewcommand\arraystretch{0.8}
\resizebox{\linewidth}{!}{

	\begin{tabular}{c|c|c|c|c|c|c}
    \toprule
    \textbf{Method}&\textbf{Metric}&\textbf{Need Training}& \textbf{Training Data} & \textbf{Data Size}& \textbf{Training Module} & \textbf{Trainig Method}\\
    \midrule
    SLEB&Perplexity&No&None&None&None&None\\
    \midrule
    ShortGPT&Cosine Similarity&No&None&None&None&None\\
    \midrule
    UIDL&Cosine Similarity&Yes&C4&164M&LoRA-Adapter&QLoRA\\
    \midrule
    LaCO&Cosine Similarity&Yes&Unpublished&1B&Full Parameters&Fine-tuning\\
    \midrule
    Shortened Llama&\makecell[c]{Taylor\\Perplexity}&Yes&\makecell[c]{SlimPajama\\Alpaca}&\makecell[c]{627B\\50k}&\makecell[c]{Full Parameters\\LoRA-Adapter}&\makecell[c]{Fine-tuning\\LoRA}\\
    \midrule
    LLM-Streamline&Cosine Similarity&Yes&SlimPajama&30k&\makecell[c]{Lightweight\\Network}& \makecell[c]{Training\\ Lightweight Network}\\
    \bottomrule
\end{tabular}
 }
	\label{tab: Layer Pruning Method}
\end{table}

\section{Related Work}
Previous pruning methods for LLMs primarily focus on pruning dense matrices~\citep{ashkboos2024slicegpt}, attention heads~\citep{michel2019sixteen, voita2019mha}, filters~\citep{mccarley2019filter, prasanna2020filter}, or hidden dimension~\citep{xia2023sheared, van2023llm}. These approaches often lead to structural irregularities, making pruned models less flexible for deployment. In contrast, layer pruning, which only alters the model's depth, is easier to deploy. Concurrent works in layer pruning alongside LLM-Streamline include LaCo~\citep{yang2024laco}, ShortGPT~\citep{men2024shortgpt}, UIDL~\citep{gromov2024unreasonable}, SLEB~\citep{song2024slebstreamliningllmsredundancy}, and Shortened Llama~\citep{kim2024shortened}. 

LaCo~\citep{yang2024laco} divides layers into groups of consecutive layers and compresses them by replacing the consecutive layers with averaged parameter weights. ShortGPT~\citep{men2024shortgpt} uses a BI score, equivalent to cosine similarity, to assess layer importance and remove less important layers. Similarly, UIDL~\citep{gromov2024unreasonable} uses angular distance, also equivalent to cosine similarity, to determine and remove less important layers, and employs QLoRA to enhance performance. SLEB~\citep{song2024slebstreamliningllmsredundancy} calculates layer importance using perplexity and discards those deemed insignificant. Shortened Llama~\citep{kim2024shortened} explores various layer selection metrics and examines the effectiveness of using continual pre-training and LoRA after pruning. The differences between these layer pruning methods and LLM-Streamline are summarized in Table \ref{tab: Layer Pruning Method}. 

Unlike traditional layer pruning methods, LLM-Streamline fundamentally differs by retraining a lightweight model to replace the pruned layers, rather than removing them directly with or without training the pruned model.
\smodel reduces both computation time and resource consumption compared to layer pruning methods (Shortened Llama, LaCo, UIDL) that necessitate retraining. Additionally, LLM-Streamline better preserves the performance of the original LLM compared to concurrent layer pruning methods.

\section{Conclusion} 
In this paper, we propose LLM-Streamline, a layer pruning-and-replacement algorithm for LLMs. We also identify shortcomings in the existing accuracy metric and propose a new metric called stability for evaluating model compression. Extensive experiments show that this layer replacement method using a lightweight network outperforms previous state-of-the-art pruning methods and demonstrates superior effectiveness and efficiency compared to concurrent layer pruning methods.

\bibliography{iclr2025_conference}

\begin{thebibliography}{54}
\providecommand{\natexlab}[1]{#1}
\providecommand{\url}[1]{\texttt{#1}}
\expandafter\ifx\csname urlstyle\endcsname\relax
  \providecommand{\doi}[1]{doi: #1}\else
  \providecommand{\doi}{doi: \begingroup \urlstyle{rm}\Url}\fi

\bibitem[Ashkboos et~al.(2024)Ashkboos, Croci, Nascimento, Hoefler, and Hensman]{ashkboos2024slicegpt}
Saleh Ashkboos, Maximilian~L Croci, Marcelo Gennari~do Nascimento, Torsten Hoefler, and James Hensman.
\newblock Slicegpt: Compress large language models by deleting rows and columns.
\newblock \emph{arXiv preprint arXiv:2401.15024}, 2024.

\bibitem[Bisk et~al.(2020)Bisk, Zellers, Gao, Choi, et~al.]{bisk2020piqa}
Yonatan Bisk, Rowan Zellers, Jianfeng Gao, Yejin Choi, et~al.
\newblock Piqa: Reasoning about physical commonsense in natural language.
\newblock In \emph{Proceedings of the AAAI conference on artificial intelligence}, volume~34, pp.\  7432--7439, 2020.

\bibitem[Chen et~al.(2023)Chen, Ding, Yadav, Zharkov, and Liang]{chen2023lorashear}
Tianyi Chen, Tianyu Ding, Badal Yadav, Ilya Zharkov, and Luming Liang.
\newblock Lorashear: Efficient large language model structured pruning and knowledge recovery.
\newblock \emph{arXiv preprint arXiv:2310.18356}, 2023.

\bibitem[Chen et~al.(2020)Chen, Kornblith, Norouzi, and Hinton]{chen2020simple}
Ting Chen, Simon Kornblith, Mohammad Norouzi, and Geoffrey Hinton.
\newblock A simple framework for contrastive learning of visual representations.
\newblock In \emph{International conference on machine learning}, pp.\  1597--1607. PMLR, 2020.

\bibitem[Chen \& He(2021)Chen and He]{chen2021exploring}
Xinlei Chen and Kaiming He.
\newblock Exploring simple siamese representation learning.
\newblock In \emph{Proceedings of the IEEE/CVF conference on computer vision and pattern recognition}, pp.\  15750--15758, 2021.

\bibitem[Clark et~al.(2019)Clark, Lee, Chang, Kwiatkowski, Collins, and Toutanova]{clark2019boolq}
Christopher Clark, Kenton Lee, Ming-Wei Chang, Tom Kwiatkowski, Michael Collins, and Kristina Toutanova.
\newblock Boolq: Exploring the surprising difficulty of natural yes/no questions.
\newblock \emph{arXiv preprint arXiv:1905.10044}, 2019.

\bibitem[Cobbe et~al.(2021)Cobbe, Kosaraju, Bavarian, Chen, Jun, Kaiser, Plappert, Tworek, Hilton, Nakano, et~al.]{cobbe2021gsm8k}
Karl Cobbe, Vineet Kosaraju, Mohammad Bavarian, Mark Chen, Heewoo Jun, Lukasz Kaiser, Matthias Plappert, Jerry Tworek, Jacob Hilton, Reiichiro Nakano, et~al.
\newblock Training verifiers to solve math word problems.
\newblock \emph{arXiv preprint arXiv:2110.14168}, 2021.

\bibitem[Contributors(2023)]{contributors2023opencompass}
OpenCompass Contributors.
\newblock Opencompass: A universal evaluation platform for foundation models.
\newblock \emph{GitHub repository}, 2023.

\bibitem[Das et~al.(2023)Das, Ma, and Shen]{das2023beyond}
Rocktim~Jyoti Das, Liqun Ma, and Zhiqiang Shen.
\newblock Beyond size: How gradients shape pruning decisions in large language models.
\newblock \emph{arXiv preprint arXiv:2311.04902}, 2023.

\bibitem[Dettmers et~al.(2022)Dettmers, Lewis, Belkada, and Zettlemoyer]{dettmers2022int8}
Tim Dettmers, Mike Lewis, Younes Belkada, and Luke Zettlemoyer.
\newblock Gpt3. int8 (): 8-bit matrix multiplication for transformers at scale.
\newblock \emph{Advances in Neural Information Processing Systems}, 35:\penalty0 30318--30332, 2022.

\bibitem[Dettmers et~al.(2023)Dettmers, Pagnoni, Holtzman, and Zettlemoyer]{dettmers2023qlora}
Tim Dettmers, Artidoro Pagnoni, Ari Holtzman, and Luke Zettlemoyer.
\newblock Qlora: Efficient finetuning of quantized llms, 2023.

\bibitem[Frantar \& Alistarh(2023)Frantar and Alistarh]{frantar2023sparsegpt}
Elias Frantar and Dan Alistarh.
\newblock Sparsegpt: Massive language models can be accurately pruned in one-shot.
\newblock In \emph{International Conference on Machine Learning}, pp.\  10323--10337. PMLR, 2023.

\bibitem[Geva et~al.(2021)Geva, Khashabi, Segal, Khot, Roth, and Berant]{geva2021did}
Mor Geva, Daniel Khashabi, Elad Segal, Tushar Khot, Dan Roth, and Jonathan Berant.
\newblock Did aristotle use a laptop? a question answering benchmark with implicit reasoning strategies.
\newblock \emph{Transactions of the Association for Computational Linguistics}, 9:\penalty0 346--361, 2021.

\bibitem[Gholami et~al.(2022)Gholami, Kim, Dong, Yao, Mahoney, and Keutzer]{gholami2022survey}
Amir Gholami, Sehoon Kim, Zhen Dong, Zhewei Yao, Michael~W Mahoney, and Kurt Keutzer.
\newblock A survey of quantization methods for efficient neural network inference.
\newblock In \emph{Low-Power Computer Vision}, pp.\  291--326. Chapman and Hall/CRC, 2022.

\bibitem[Gou et~al.(2021)Gou, Yu, Maybank, and Tao]{gou2021knowledge}
Jianping Gou, Baosheng Yu, Stephen~J Maybank, and Dacheng Tao.
\newblock Knowledge distillation: A survey.
\newblock \emph{International Journal of Computer Vision}, 129\penalty0 (6):\penalty0 1789--1819, 2021.

\bibitem[Gromov et~al.(2024)Gromov, Tirumala, Shapourian, Glorioso, and Roberts]{gromov2024unreasonable}
Andrey Gromov, Kushal Tirumala, Hassan Shapourian, Paolo Glorioso, and Daniel~A Roberts.
\newblock The unreasonable ineffectiveness of the deeper layers.
\newblock \emph{arXiv preprint arXiv:2403.17887}, 2024.

\bibitem[Hendrycks et~al.(2020)Hendrycks, Burns, Basart, Zou, Mazeika, Song, and Steinhardt]{hendrycks2020mmlu}
Dan Hendrycks, Collin Burns, Steven Basart, Andy Zou, Mantas Mazeika, Dawn Song, and Jacob Steinhardt.
\newblock Measuring massive multitask language understanding.
\newblock \emph{arXiv preprint arXiv:2009.03300}, 2020.

\bibitem[Hinton et~al.(2015)Hinton, Vinyals, and Dean]{hinton2015distilling}
Geoffrey Hinton, Oriol Vinyals, and Jeff Dean.
\newblock Distilling the knowledge in a neural network.
\newblock \emph{arXiv preprint arXiv:1503.02531}, 2015.

\bibitem[Ho et~al.(2022)Ho, Schmid, and Yun]{ho2022distill}
Namgyu Ho, Laura Schmid, and Se-Young Yun.
\newblock Large language models are reasoning teachers.
\newblock \emph{arXiv preprint arXiv:2212.10071}, 2022.

\bibitem[Hu et~al.(2021)Hu, Shen, Wallis, Allen-Zhu, Li, Wang, Wang, and Chen]{hu2021lora}
Edward~J. Hu, Yelong Shen, Phillip Wallis, Zeyuan Allen-Zhu, Yuanzhi Li, Shean Wang, Lu~Wang, and Weizhu Chen.
\newblock Lora: Low-rank adaptation of large language models, 2021.

\bibitem[Hu et~al.(2024)Hu, Zhang, Zhao, Zhao, Chen, Li, and Chen]{sp3-hu2024}
Yuxuan Hu, Jing Zhang, Zhe Zhao, Chen Zhao, Xiaodong Chen, Cuiping Li, and Hong Chen.
\newblock $\rm sp^3$: Enhancing structured pruning via pca projection, 2024.
\newblock URL \url{https://arxiv.org/abs/2308.16475}.

\bibitem[Huang et~al.(2022)Huang, Chen, Yu, and McKeown]{huang2022distill}
Yukun Huang, Yanda Chen, Zhou Yu, and Kathleen McKeown.
\newblock In-context learning distillation: Transferring few-shot learning ability of pre-trained language models.
\newblock \emph{arXiv preprint arXiv:2212.10670}, 2022.

\bibitem[Kim et~al.(2024)Kim, Kim, Kim, Castells, Choi, Shin, and Song]{kim2024shortened}
Bo-Kyeong Kim, Geonmin Kim, Tae-Ho Kim, Thibault Castells, Shinkook Choi, Junho Shin, and Hyoung-Kyu Song.
\newblock Shortened llama: A simple depth pruning for large language models.
\newblock \emph{arXiv preprint arXiv:2402.02834}, 2024.

\bibitem[Lai et~al.(2017)Lai, Xie, Liu, Yang, and Hovy]{lai2017race}
Guokun Lai, Qizhe Xie, Hanxiao Liu, Yiming Yang, and Eduard Hovy.
\newblock Race: Large-scale reading comprehension dataset from examinations.
\newblock \emph{arXiv preprint arXiv:1704.04683}, 2017.

\bibitem[Levesque et~al.(2012)Levesque, Davis, and Morgenstern]{levesque2012winograd}
Hector Levesque, Ernest Davis, and Leora Morgenstern.
\newblock The winograd schema challenge.
\newblock In \emph{Thirteenth international conference on the principles of knowledge representation and reasoning}, 2012.

\bibitem[Li et~al.(2023)Li, Zhang, Koto, Yang, Zhao, Gong, Duan, and Baldwin]{li2023cmmlu}
Haonan Li, Yixuan Zhang, Fajri Koto, Yifei Yang, Hai Zhao, Yeyun Gong, Nan Duan, and Timothy Baldwin.
\newblock Cmmlu: Measuring massive multitask language understanding in chinese.
\newblock \emph{arXiv preprint arXiv:2306.09212}, 2023.

\bibitem[Li et~al.(2022)Li, Chen, Shen, Chen, Zhang, Li, Wang, Qian, Peng, Mao, et~al.]{li2022apidistill}
Shiyang Li, Jianshu Chen, Yelong Shen, Zhiyu Chen, Xinlu Zhang, Zekun Li, Hong Wang, Jing Qian, Baolin Peng, Yi~Mao, et~al.
\newblock Explanations from large language models make small reasoners better.
\newblock \emph{arXiv preprint arXiv:2210.06726}, 2022.

\bibitem[Li et~al.(2024)Li, Ning, Wang, Liu, Shi, Yan, Dai, Yang, and Wang]{li2024evaluating}
Shiyao Li, Xuefei Ning, Luning Wang, Tengxuan Liu, Xiangsheng Shi, Shengen Yan, Guohao Dai, Huazhong Yang, and Yu~Wang.
\newblock Evaluating quantized large language models, 2024.

\bibitem[Liu et~al.(2021)Liu, Wang, Han, Zhang, Ma, and Gao]{liu2021quan}
Zhenhua Liu, Yunhe Wang, Kai Han, Wei Zhang, Siwei Ma, and Wen Gao.
\newblock Post-training quantization for vision transformer.
\newblock \emph{Advances in Neural Information Processing Systems}, 34:\penalty0 28092--28103, 2021.

\bibitem[Liu et~al.(2023)Liu, Wang, Dao, Zhou, Yuan, Song, Shrivastava, Zhang, Tian, Re, et~al.]{liu2023deja}
Zichang Liu, Jue Wang, Tri Dao, Tianyi Zhou, Binhang Yuan, Zhao Song, Anshumali Shrivastava, Ce~Zhang, Yuandong Tian, Christopher Re, et~al.
\newblock Deja vu: Contextual sparsity for efficient llms at inference time.
\newblock In \emph{International Conference on Machine Learning}, pp.\  22137--22176. PMLR, 2023.

\bibitem[Louizos et~al.(2017)Louizos, Welling, and Kingma]{louizos2017l0}
Christos Louizos, Max Welling, and Diederik~P Kingma.
\newblock Learning sparse neural networks through $ l\_0 $ regularization.
\newblock \emph{arXiv preprint arXiv:1712.01312}, 2017.

\bibitem[Ma et~al.(2023)Ma, Fang, and Wang]{ma2023llmpruner}
Xinyin Ma, Gongfan Fang, and Xinchao Wang.
\newblock Llm-pruner: On the structural pruning of large language models.
\newblock \emph{Advances in neural information processing systems}, 36:\penalty0 21702--21720, 2023.

\bibitem[McCarley et~al.(2019)McCarley, Chakravarti, and Sil]{mccarley2019filter}
JS~McCarley, Rishav Chakravarti, and Avirup Sil.
\newblock Structured pruning of a bert-based question answering model.
\newblock \emph{arXiv preprint arXiv:1910.06360}, 2019.

\bibitem[Men et~al.(2024)Men, Xu, Zhang, Wang, Lin, Lu, Han, and Chen]{men2024shortgpt}
Xin Men, Mingyu Xu, Qingyu Zhang, Bingning Wang, Hongyu Lin, Yaojie Lu, Xianpei Han, and Weipeng Chen.
\newblock Shortgpt: Layers in large language models are more redundant than you expect.
\newblock \emph{arXiv preprint arXiv:2403.03853}, 2024.

\bibitem[Michel et~al.(2019)Michel, Levy, and Neubig]{michel2019sixteen}
Paul Michel, Omer Levy, and Graham Neubig.
\newblock Are sixteen heads really better than one?
\newblock \emph{Advances in neural information processing systems}, 32, 2019.

\bibitem[Narayan et~al.(2018)Narayan, Cohen, and Lapata]{narayan2018xsum}
Shashi Narayan, Shay~B Cohen, and Mirella Lapata.
\newblock Don't give me the details, just the summary! topic-aware convolutional neural networks for extreme summarization.
\newblock \emph{arXiv preprint arXiv:1808.08745}, 2018.

\bibitem[Prasanna et~al.(2020)Prasanna, Rogers, and Rumshisky]{prasanna2020filter}
Sai Prasanna, Anna Rogers, and Anna Rumshisky.
\newblock When bert plays the lottery, all tickets are winning.
\newblock \emph{arXiv preprint arXiv:2005.00561}, 2020.

\bibitem[Reimers(2019)]{reimers2019sentence}
N~Reimers.
\newblock Sentence-bert: Sentence embeddings using siamese bert-networks.
\newblock \emph{arXiv preprint arXiv:1908.10084}, 2019.

\bibitem[Song et~al.(2024)Song, Oh, Kim, Kim, Kim, and Kim]{song2024slebstreamliningllmsredundancy}
Jiwon Song, Kyungseok Oh, Taesu Kim, Hyungjun Kim, Yulhwa Kim, and Jae-Joon Kim.
\newblock Sleb: Streamlining llms through redundancy verification and elimination of transformer blocks, 2024.
\newblock URL \url{https://arxiv.org/abs/2402.09025}.

\bibitem[Sun et~al.(2020)Sun, Yu, Yu, and Cardie]{sun2020C3}
Kai Sun, Dian Yu, Dong Yu, and Claire Cardie.
\newblock Investigating prior knowledge for challenging chinese machine reading comprehension.
\newblock \emph{Transactions of the Association for Computational Linguistics}, 8:\penalty0 141--155, 2020.

\bibitem[Sun et~al.(2023)Sun, Liu, Bair, and Kolter]{sun2023simple}
Mingjie Sun, Zhuang Liu, Anna Bair, and J~Zico Kolter.
\newblock A simple and effective pruning approach for large language models.
\newblock \emph{arXiv preprint arXiv:2306.11695}, 2023.

\bibitem[Talmor et~al.(2018)Talmor, Herzig, Lourie, and Berant]{talmor2018commonsenseqa}
Alon Talmor, Jonathan Herzig, Nicholas Lourie, and Jonathan Berant.
\newblock Commonsenseqa: A question answering challenge targeting commonsense knowledge.
\newblock \emph{arXiv preprint arXiv:1811.00937}, 2018.

\bibitem[Touvron et~al.(2023)Touvron, Martin, Stone, Albert, Almahairi, Babaei, Bashlykov, Batra, Bhargava, Bhosale, et~al.]{touvron2023llama}
Hugo Touvron, Louis Martin, Kevin Stone, Peter Albert, Amjad Almahairi, Yasmine Babaei, Nikolay Bashlykov, Soumya Batra, Prajjwal Bhargava, Shruti Bhosale, et~al.
\newblock Llama 2: Open foundation and fine-tuned chat models.
\newblock \emph{arXiv preprint arXiv:2307.09288}, 2023.

\bibitem[van~der Ouderaa et~al.(2023)van~der Ouderaa, Nagel, Van~Baalen, Asano, and Blankevoort]{van2023llm}
Tycho~FA van~der Ouderaa, Markus Nagel, Mart Van~Baalen, Yuki~M Asano, and Tijmen Blankevoort.
\newblock The llm surgeon.
\newblock \emph{arXiv preprint arXiv:2312.17244}, 2023.

\bibitem[Vaswani et~al.(2017)Vaswani, Shazeer, Parmar, Uszkoreit, Jones, Gomez, Kaiser, and Polosukhin]{vaswani2017attention}
Ashish Vaswani, Noam Shazeer, Niki Parmar, Jakob Uszkoreit, Llion Jones, Aidan~N Gomez, {\L}ukasz Kaiser, and Illia Polosukhin.
\newblock Attention is all you need.
\newblock \emph{Advances in neural information processing systems}, 30, 2017.

\bibitem[Voita et~al.(2019)Voita, Talbot, Moiseev, Sennrich, and Titov]{voita2019mha}
Elena Voita, David Talbot, Fedor Moiseev, Rico Sennrich, and Ivan Titov.
\newblock Analyzing multi-head self-attention: Specialized heads do the heavy lifting, the rest can be pruned, 2019.

\bibitem[Wang et~al.(2024)Wang, Chen, Luo, Long, Lin, Zhang, Lin, Cai, and He]{wang2024survey}
Wenxiao Wang, Wei Chen, Yicong Luo, Yongliu Long, Zhengkai Lin, Liye Zhang, Binbin Lin, Deng Cai, and Xiaofei He.
\newblock Model compression and efficient inference for large language models: A survey, 2024.

\bibitem[Xia et~al.(2023)Xia, Gao, Zeng, and Chen]{xia2023sheared}
Mengzhou Xia, Tianyu Gao, Zhiyuan Zeng, and Danqi Chen.
\newblock Sheared llama: Accelerating language model pre-training via structured pruning.
\newblock \emph{arXiv preprint arXiv:2310.06694}, 2023.

\bibitem[Xu et~al.(2020)Xu, Hu, Zhang, Li, Cao, Li, Xu, Sun, Yu, Yu, et~al.]{xu2020clue}
Liang Xu, Hai Hu, Xuanwei Zhang, Lu~Li, Chenjie Cao, Yudong Li, Yechen Xu, Kai Sun, Dian Yu, Cong Yu, et~al.
\newblock Clue: A chinese language understanding evaluation benchmark.
\newblock \emph{arXiv preprint arXiv:2004.05986}, 2020.

\bibitem[Yang et~al.(2024)Yang, Cao, and Zhao]{yang2024laco}
Yifei Yang, Zouying Cao, and Hai Zhao.
\newblock Laco: Large language model pruning via layer collapse.
\newblock \emph{arXiv preprint arXiv:2402.11187}, 2024.

\bibitem[Zellers et~al.(2019)Zellers, Holtzman, Bisk, Farhadi, and Choi]{zellers2019hellaswag}
Rowan Zellers, Ari Holtzman, Yonatan Bisk, Ali Farhadi, and Yejin Choi.
\newblock Hellaswag: Can a machine really finish your sentence?
\newblock \emph{arXiv preprint arXiv:1905.07830}, 2019.

\bibitem[Zhang et~al.(2022)Zhang, Roller, Goyal, Artetxe, Chen, Chen, Dewan, Diab, Li, Lin, et~al.]{zhang2022opt}
Susan Zhang, Stephen Roller, Naman Goyal, Mikel Artetxe, Moya Chen, Shuohui Chen, Christopher Dewan, Mona Diab, Xian Li, Xi~Victoria Lin, et~al.
\newblock Opt: Open pre-trained transformer language models.
\newblock \emph{arXiv preprint arXiv:2205.01068}, 2022.

\bibitem[Zheng et~al.(2019)Zheng, Huang, and Sun]{zheng2019chid}
Chujie Zheng, Minlie Huang, and Aixin Sun.
\newblock Chid: A large-scale chinese idiom dataset for cloze test.
\newblock \emph{arXiv preprint arXiv:1906.01265}, 2019.

\bibitem[Zhu et~al.(2023)Zhu, Li, Liu, Ma, and Wang]{zhu2023survey}
Xunyu Zhu, Jian Li, Yong Liu, Can Ma, and Weiping Wang.
\newblock A survey on model compression for large language models, 2023.

\end{thebibliography}
\bibliographystyle{iclr2025_conference}

\newpage
\appendix

%\section{Comparison of Cosine Similarity and other similarity Metric\label{sec: similarity metric}} 

\section{Comparison of Cosine Similarity and Perplexity\label{sec: Comparison}}
To demonstrate the sensitivity of perplexity, referencing the SLEB~\citep{song2024slebstreamliningllmsredundancy}, we prune the Llama2-7B model with different pre-training datasets, including SlimPajama, C4 and wikitext. The experimental results are presented in Table \ref{tab:Pruning results of comparison}. When pruning with cosine similarity, the layers pruned are consistent across different datasets, whereas when pruning with perplexity, the layers vary, indicating the sensitivity of perplexity. In addition, we evaluate the model after pruning with the SlimPajama dataset, and the experimental results are shown in Table \ref{tab:detailed result of comparison}. This indicates that the model pruned with perplexity shows lower perplexity on the dataset used for pruning, but performs worse on downstream tasks.
%We present detailed results of using perplexity and cosine similarity for pruning in the Table \ref{tab:detailed result of comparison}. We select several benchmarks where perplexity performs poorly as the pruning metric, including BoolQ, CoQA, HeSW, Race-H, and Race-M, and calculate the average perplexity of the pruned model on these benchmarks. As shown in the Table \ref{}, we find that when using cosine similarity as the pruning metric, the perplexity of the pruned model on these benchmarks is significantly lower than when using perplexity as the metric.

\begin{table}[H]
\caption{Pruned layers using perplexity and cosine similarity for pruning.}
\centering
\renewcommand\arraystretch{1.2}
\resizebox{0.7\linewidth}{!}{
	\begin{tabular}{c|c|cc}
 \toprule
\multirow{2}{*}{LLM} & \multirow{2}{*}{Dataset}    & \multicolumn{2}{c}{Pruned Layers}           \\
 & & \textbf{Perplexity}      & \textbf{Cosine Similarity} \\
 \midrule
 \multirow{3}{*}{Llama2-7B}
                           & SlimPajama & 9,10,11,12,21,23,25,27   & 22,23,24,25,26,27,28,29 \\
                
                           & wikitext & 9,10,11,12,21,23,24,27
                           & 22,23,24,25,26,27,28,29 \\

                           & C4 & 8,9,11,12,22,23,24,25    & 22,23,24,25,26,27,28,29 \\
                           %& BoolQ      & 8,12,14,15,23,24,25,27    & 23,24,25,26,27,28,29,30 \\
                           %& PIQA       & 10,11,12,17,18,21,25,26 & 23,24,25,26,27,28,29,30 \\
                           %& HeSW       & 12,13,14,15,22,23,25,27 & 23,24,25,26,27,28,29,30 \\
                           %& C3         & 9,10,11,12,21,23,25,27 & 23,24,25,26,27,28,29,30 \\
                           \bottomrule
\end{tabular}
 }
	\label{tab:Pruning results of comparison}
\end{table}

\begin{table}[H]
\caption{Detailed results of accuracy of using perplexity and cosine similarity for pruning. ``Perplexity*'' refers to the Perplexity of the pruned model on SlimPajama. Using perplexity as the metric can be considered as SLEB, while using cosine similarity as the metric can be considered as a variant of our approach, i.e., Ours (None)(details in Section~\ref{sec: setup}).}
	\small
	\centering
\renewcommand\arraystretch{1.4}
\resizebox{1\linewidth}{!}{

	\begin{tabular}{c|c|c|ccccccccccccccc|cc}
		\toprule
  \multirow{2}{*}{LLM} & \multirow{2}{*}{Metric} &\multirow{2}{*}{Perplexity*} & \multicolumn{15}{c|}{Benchmarks}& \multirow{2}{*}{Average} &\multirow{2}{*}{RP} \ \\  
		& & & C3 & CMNLI&CHID&BoolQ&WSC&CoQA&HeSW&PIQA&Race-M&Race-H &MMLU & CMMLU&Xsum &GSM8k &StrategyQA  \\
		\midrule
        \multirow{3}{*}{Llama2-7B} 
  & Dense&6.23   &43.8 & 33.0 & 41.6 & 70.8 & 37.5 & 66.7 & 71.3 & 78.1 & 33.1 & 35.5 & 46.8 & 31.8 &19.4&16.5&60.2& 45.7 & 100.0\\
             & Cosine Similarity &19.7
&\textbf{40.2} & \textbf{34.4} & 21.5 & \textbf{67.3} & \textbf{40.4} & \textbf{51.7} & \textbf{59.7} & 69.0 & \textbf{35.5} & \textbf{34.7} & \textbf{44.6} & \textbf{28.9} &14.8&\textbf{1.97}&\textbf{41.8}& \textbf{39.1} & \textbf{85.6}   \\
             & Perplexity  &\textbf{12.1}&37.6 & 33.0 & \textbf{34.2} & 61.7 & 36.5 & 47.3 & 56.5 & \textbf{71.4} & 22.1 & 21.6 & 25.9 & 24.8 &\textbf{17.1}&1.74&33.2 & 35.0 & 76.6\\
                   
		\bottomrule
	\end{tabular}
 }
	\label{tab:detailed result of comparison}
\end{table}

\section{Data Distribution\label{sec: proportion}}

We extract the training data from different domains based on the data distribution strategy proposed in Sheared-LLaMa~~\citep{xia2023sheared}. The detailed data distribution is shown in Table~\ref{tab: proportion}. 

\begin{table}[H]
\caption{The proportion of different domains randomly selected from the SlimPajama-6B dataset.
}
\centering
\renewcommand\arraystretch{1.2}
\resizebox{0.8\linewidth}{!}{
\begin{tabular}{c|cccccccc|}
\toprule
 & \textbf{CC} & \textbf{GitHub} & \textbf{Book} & \textbf{StackExchange} & \textbf{Wiki} & \textbf{ArXiv} &\textbf{C4} \\ 
\midrule
SlimPajama-6B & 54.1\% & 4.2\% & 3.7\% & 2.8\% & 3.1\% & 3.4\% & 28.7\%\\ 
Ours      & 36.1\% & 0.8\% & 9.1\% & 1.0\% & 3.1\% & 0.7\% & 49.2\% \\
\bottomrule
\end{tabular}
}
\label{tab: proportion}
\end{table}

% \section{The proportion of different domains randomly selected from the SlimPajama-6B dataset \label{sec: proportion}}
\section{Training Implementation Details\label{sec: Training Details}}
\subsection{Lightweight Network Training Details\label{sec: lightweight network training details}}
For both the FFN structure and the SwiGLU structure, the learning rate is set to 1e-3 and the weight decay is 1e-4. For the Transformer layer, the learning rate is set to 1e-5 and the weight decay is 1e-3. The model is trained using a batch size of 32 over 20 epochs. On a single A800 GPU, the training duration for the lightweight network is approximate 5 hours (for the Transformer layer).
\subsection{Post Training Details \label{sec: post-training details}}
For layer replacement, in order to have a fairer comparison with LoRA, we conduct one epoch of post-training with a learning rate of 5e-5, a weight decay of 1e-3, and a batch size of 32. This process takes less than an hour on a single A800 GPU. For LoRA, the model is trained one epochs with a learning rate of 1e-4, a weight decay of 1e-3, and a batch size of 32. Since the amount of training data used is ten times that of layer replacement, it take approximately 10 hours to complete the training on a single A800 GPU.

\section{Inference speed comparison \label{sec: speed time}}
As shown in the Table~\ref{tab:speed}, we evaluate the inference speed of various pruning methods at similar pruning ratio when generating sequences of length 128. The results indicate that the acceleration effect of LLM-Streamline is slightly inferior to that of SliceGPT and LLM-Pruner.

\begin{table}[H]
\caption{The inference speed of models pruned using different methods.
}
\centering
\renewcommand\arraystretch{1.2}
\resizebox{0.8\linewidth}{!}{
\begin{tabular}{c|ccccccc|}
\toprule
 Llama2-7B & Dense & LLM-Pruner & SliceGPT & Ours(None) & Ours(FFN) & Ours(Layer) \\ 
\midrule
Pruning Ratio (\%) &0.00 &24.8 &25.4 &24.0 &25.0 &24.0 \\
Inference Speed (tokens/s) & 19.87 &25.91 &27.20 &25.68 &25.88 &25.68 \\ 
\bottomrule
\end{tabular}
}
\label{tab:speed}
\end{table}

% \section{MLP Training Implementation Details}
% As mentioned in Section 4.3, for OPT-1.3B and OPT-2.7B, we train the MLP using 3,000 randomly selected samples from the wikitext-2 dataset. The batch size is set to 512, the learning rate to 4e-4, and the number of epochs to 5, which could be completed in just 15 minutes on a single RTX 3090. For OPT-6.7B and Llama2-7B, we train the MLP using 10,000 or 20,000 randomly selected samples from the SlimPajama-6B dataset. The batch size is set to 2048, the learning rate to 1e-3, and the number of epochs to 5, requiring approximately 2.5 hours to complete the training on a single A100.

% \section{Model Structure, Number of Parameters, and Amount of Training Data}

\section{Detailed Experimental Results}
\subsection{Experimental Results of OPT-6.7B \label{sec: Result of OPT-6.7B and OPT-13B}}
We conduct experiments on OPT-6.7B. The experimental results are shown in Table~\ref{tab:opt-6.7b}, Table \ref{tab:opt-6.7b-} and Table \ref{tab:opt-6.7b+}.
The results indicate that our proposed LLM-Streamline is superior to the previous SOTA method.

\begin{table}[H]
\caption{Accuracy of different pruning methods on classification benchmarks by pruning OPT-6.7B.}
	\small
	\centering
\renewcommand\arraystretch{1.2}
\resizebox{\linewidth}{!}{

	\begin{tabular}{c|c|c|cccccccccccc|cc}
		\toprule
		\multirow{2}{*}{LLM} & \multirow{2}{*}{Method}  & \multirow{2}{*}{Ratio}& \multicolumn{12}{c|}{Benchmarks}& \multirow{2}{*}{Average} & \multirow{2}{*}{RP} \\  
		&   & & C3 & CMNLI&CHID&BoolQ&WSC&CoQA&HeSW&PIQA&Race-M&Race-H &MMLU & CMMLU& &\\
   \midrule
            \multirow{5}{*}{OPT-6.7B} 
  & Dense  & 0.00\% &38.7  &32.9   &21.6  &64.6 &41.4 &54.8 &63.3 &75.4&25.1   &25.4&24.7  &25.5  &41.1 &100
             \\
             & SliceGPT & 25.6\%&\textbf{40.0}  &31.2&\textbf{19.5}&37.9&36.5&38.2&45.6&65.8&\textbf{25.8}&\textbf{26.0}&\textbf{25.8}&24.8&34.8&84.7\\
             % & ShortGPT*  & 27.0\%&26.6  &32.6&14.5  &43.6 &38.5 &18.8 &25.7 &50.8&22.0   &22.7&24.0  &25.5  &28.8 &70.1\\
             & Ours (None)  & 24.0\%&27.6  &\textbf{32.5}&12.7  &44.8 &36.5 &20.6 &26.5 &52.1&22.1   &22.4&23.6  &\textbf{25.2}  &28.9 &70.3\\
             & Ours (FFN) & 25.0\%&\underline{37.6}  &\underline{32.1}& 18.7 &\textbf{63.7} &\underline{37.5} &\underline{41.8} &\textbf{55.9} &\underline{73.2}&22.7   &22.2&\underline{24.4}  &\underline{24.9}  &\underline{37.9} &\underline{92.2}\\
             & Ours (Layer) & 24.0\%& 36.4 &32.0&\underline{18.9}  &\underline{62.4} &\textbf{38.5} &\textbf{45.1} &\underline{54.3} &\textbf{74.0}&\underline{23.6}   &\underline{24.2}& 24.3 &\textbf{25.2}  &\textbf{38.2} &\textbf{92.9}\\          
		\bottomrule
	\end{tabular}
 }
	\label{tab:opt-6.7b}
\end{table}

\begin{table}[H]
\caption{Stability of different pruning methods on classification benchmarks by pruning OPT-6.7B.}
	\small
	\centering
\renewcommand\arraystretch{1.2}
\resizebox{\linewidth}{!}{

	\begin{tabular}{c|c|c|cccccccccccc|c}
		\toprule
		\multirow{2}{*}{LLM} & \multirow{2}{*}{Method}  & \multirow{2}{*}{Ratio}& \multicolumn{12}{c|}{Benchmarks}& \multirow{2}{*}{Average} \\  
		&   & & C3 & CMNLI&CHID&BoolQ&WSC&CoQA&HeSW&PIQA&Race-M&Race-H &MMLU & CMMLU& \\
    \midrule
            \multirow{4}{*}{OPT-6.7B} 
            & SliceGPT & 25.6\% &66.4  &\underline{39.8}&73.1  &30.1 &79.4  &75.2 &73.7 &82.5 &\underline{72.5} &\underline{69.1} &68.9   &\underline{67.2}    &66.5 \\
            % & ShortGPT* & 27.0\% &55.2  &73.5&72.3  &38.5 &89.4  &47.0 &48.2 &55.2 &61.3 &60.5&61.0   &62.7    &60.4 \\
            & Ours (None) & 24.0\% &56.2  &\textbf{62.5}& 71.4 &41.5 &87.5  &48.6 &50.5 &57.1 &63.4 &62.2&62.3   &62.7    &60.5 \\
            & Ours (FFN) & 25.0\% &\textbf{74.1}  &36.1&\textbf{77.4}  &\textbf{74.2} &\underline{90.4}  &\underline{82.4} &\textbf{88.4} &\textbf{92.0} &68.3 &63.9&\underline{74.7}   &65.0    & \underline{73.9}\\
            & Ours (Layer) & 24.0\% &\underline{72.1} &35.5&\underline{76.1} &\underline{72.1} &\textbf{91.4}  &\textbf{83.6} &\underline{87.0} &\underline{91.6} &\textbf{72.9} &\textbf{71.2}&\textbf{78.1}   &\textbf{70.1}    &\textbf{75.1} \\
		\bottomrule
	\end{tabular}
 }
	\label{tab:opt-6.7b-}
\end{table}

\begin{table}[H]
\caption{Accuracy of different pruning methods on generation benchmarks by pruning OPT-6.7B.}
\small
\centering
\renewcommand\arraystretch{1.2}
\resizebox{0.75\linewidth}{!}{
\begin{tabular}{c|c|c|ccc|cc}
    \toprule

    \multirow{2}{*}{LLM} & \multirow{2}{*}{Method}  & \multirow{2}{*}{Ratio}& \multicolumn{3}{c|}{Benchmarks}& \multirow{2}{*}{Average} & \multirow{2}{*}{RP} \\  
    &   & & Xsum & GSM8K & StrategyQA& &\\
    \midrule
    \multirow{5}{*}{OPT-6.7B} 
    & Dense  & 0.00\% &13.4 &2.2 &54.3 &23.3 &100.0
    \\
    & SliceGPT & 25.6\% &\underline{14.9} &\textbf{2.5} &40.8 &19.4 &83.3\\
    % & ShortGPT* & 27.0\% &3.7 &0.2 &35.3 &13.1 &56.2\\
    & Ours (None) & 24.0\% &4.9 &0 &0 &1.6 &6.87\\
    & Ours (FFN) & 25.0\% &14.8 &\underline{0.8} &\underline{43.6} &\underline{19.7} &\underline{84.5}\\
    & Ours (Layer) & 24.0\% &\textbf{18.4} &\textbf{2.5} &\textbf{44.4} &\textbf{21.8} &\textbf{93.6}\\

    \bottomrule
\end{tabular}
}
\label{tab:opt-6.7b+}
\end{table}

\subsection{Experimental Results of Baichuan-7B and Baichuan-13B \label{sec: Result of Baichuan-7B and Baichuan-13B}}
We conduct experiments on Baichuan-7B and Baichuan-13B. The experimental results are shown in Table~\ref{tab:accuracy of Baichuan}, Table~\ref{tab:baichuan_stability} and Table~\ref{tab:baichuan_gen}.
The results indicate that our proposed LLM-Streamline is superior to the previous SOTA method.

\begin{table}[H]
\caption{Accuracy of different pruning methods on classification benchmarks by pruning Baichuan-7B and Baichuan-13B.}
	\small
	\centering
\renewcommand\arraystretch{1.2}
\resizebox{\linewidth}{!}{

	\begin{tabular}{c|c|c|cccccccccccc|cc}
		\toprule
		\multirow{2}{*}{LLM} & \multirow{2}{*}{Method}  & \multirow{2}{*}{Ratio}& \multicolumn{12}{c|}{Benchmarks}& \multirow{2}{*}{Average} & \multirow{2}{*}{RP} \\  
		&   & & C3 & CMNLI&CHID&BoolQ&WSC&CoQA&HeSW&PIQA&Race-M&Race-H &MMLU & CMMLU& &\\
		\midrule
		\multirow{5}{*}{Baichuan-7B} 
  & Dense  & 0.00\% &55.8&35.3&91.3&61.4&39.4&58.4&65.3&77.6&29.5&30.4&43.7&43.8&52.7&100
		\\
	& LLMPruner & 24.2\% &43.7&33.9&65.9&40.5&\textbf{36.5}&48.2&52.2&68.1&22.6&22.0&24.2&25.3&40.2&76.3\\
        & Ours (None)  & 24.2\% &33.2&32.7&25.8&\textbf{60.8}&\textbf{36.5}&36.0&34.6&58.7&22.1&21.5&25.7&38.8&35.5&67.4\\
        & Ours (FFN)  & 25.1\% &\underline{53.1}&\textbf{36.3}&\underline{69.4}&\underline{53.1}&\textbf{36.5}&\underline{48.7}&\underline{53.2}&\textbf{69.4}&\textbf{23.2}&\textbf{24.5}&\underline{37.7}&\underline{39.1}&\underline{45.4}&\underline{86.1}\\
        & Ours (Layer) & 24.2\% &\textbf{55.0}&\underline{36.0}&\textbf{77.4}&48.1&\textbf{36.5}&\textbf{49.8}&\textbf{54.3}&\underline{69.0}&\underline{22.9}&\underline{23.8}&\textbf{39.8}&\textbf{41.1}&\textbf{46.1}&\textbf{87.5}\\
            
            \midrule
            \multirow{4}{*}{Baichuan-13B} 
  & Dense  & 0.00\% &61.5 &36.4&91.5&65.8&49.0&64.2&69.1&78.2&48.1&46.0&54.8&55.3&60.0 &100
		\\
        & Ours (None)  & 24.7\% &48.8&34.8&50.2&\underline{62.2}&\textbf{40.4}&46.4&56.7&68.2&\textbf{30.6}&\underline{27.7}&\underline{52.9}&\underline{55.1}&47.8&79.7\\
        & Ours (FFN)  & 25.5\% &\underline{58.3}&\underline{35.1}&\underline{77.5}&\textbf{64.1}&\underline{36.5}&\underline{57.7}&\underline{58.2}&\underline{69.4}&26.5&\textbf{28.8}&\textbf{53.1}&54.3&\underline{51.6}&\underline{86.0}\\
        & Ours (Layer) & 24.7\% &\textbf{59.1}&\textbf{36.1}&\textbf{83.7}&62.0&\underline{36.5}&\textbf{58.2}&\textbf{59.4}&\textbf{71.8}&\underline{27.8}&25.0&52.3&\textbf{56.1}&\textbf{52.3}&\textbf{87.2}\\     
		\bottomrule
	\end{tabular}
 }
	\label{tab:accuracy of Baichuan}
\end{table}

\begin{table}[H]
\caption{Stability of different pruning methods on classification benchmarks by pruning Baichuan-7B and Baichuan-13B.}
	\small
	\centering
\renewcommand\arraystretch{1.2}
\resizebox{\linewidth}{!}{

	\begin{tabular}{c|c|c|cccccccccccc|c}
		\toprule
		\multirow{2}{*}{LLM} & \multirow{2}{*}{Method}  & \multirow{2}{*}{Ratio}& \multicolumn{12}{c|}{Benchmarks}& \multirow{2}{*}{Average} \\  
		&   & & C3 & CMNLI&CHID&BoolQ&WSC&CoQA&HeSW&PIQA&Race-M&Race-H &MMLU & CMMLU& \\
		\midrule
		\multirow{4}{*}{Baichuan-7B} 
    		& LLMPruner & 24.2\% &70.2&40.0&71.4&24.9&\underline{91.4}&75.3&75.9&82.4&68.8&66.8&54.7&53.6&64.6 \\
            & Ours (None) & 24.2\% &55.5&45.1&31.3&\textbf{76.8}&\textbf{93.3}&64.7&50.1&61.0&68.9&65.9&74.0&60.8&62.3\\
            & Ours (FFN) & 25.1\% &\underline{84.1}&\underline{77.8}&\underline{75.6}&\underline{65.4}&\textbf{93.3}&\underline{76.5}&\textbf{83.1}&\textbf{84.3}&\textbf{77.1}&\underline{70.3}&\underline{74.2}&\underline{74.6}& \textbf{78.0}\\
            & Ours (Layer) & 24.2\% &\textbf{86.3}&\textbf{79.3}&\textbf{82.3}&40.7&\textbf{93.3}&\textbf{77.6}&\underline{81.3}&\underline{83.5}&\underline{75.8}&\textbf{71.0}&\textbf{75.9}&\textbf{75.6}&\underline{76.9} \\
            
            \midrule
            \multirow{3}{*}{Baichuan-13B} 
            & Ours (None) & 24.7\% &67.2&75.8&54.1&66.6&\textbf{51.0}&74.9&74.6&78.5&\textbf{52.3}&\underline{55.2}&82.7&\underline{89.3}&68.5\\
            & Ours (FFN) & 25.5\% &\underline{85.7}&\underline{87.3}&\underline{82.1}&\textbf{81.3}&\underline{43.3}&\underline{81.2}&\underline{81.7}&\underline{79.8}&46.6&\textbf{61.1}&\textbf{84.3}&83.7&\underline{74.8} \\
            & Ours (Layer) & 24.7\% &\textbf{88.6}&\textbf{92.7}&\textbf{89.3}&\underline{72.9}&\underline{43.3}&\textbf{83.5}&\textbf{86.1}&\textbf{88.2}&\underline{49.2}&52.1&\underline{83.4}&\textbf{90.4}&\textbf{76.6} \\
            \bottomrule
	\end{tabular}
 }
	\label{tab:baichuan_stability}
\end{table}

\begin{table}[H]
\caption{Accuracy of different pruning methods on generation benchmarks by pruning Baichuan-7B and Baichuan-13B.}
\small
\centering
\renewcommand\arraystretch{1.2}
\resizebox{0.65\linewidth}{!}{
\begin{tabular}{c|c|c|ccc|cc}
    \toprule

    \multirow{2}{*}{LLM} & \multirow{2}{*}{Method}  & \multirow{2}{*}{Ratio}& \multicolumn{3}{c|}{Benchmarks}& \multirow{2}{*}{Average} & \multirow{2}{*}{RP} \\  
    &   & & Xsum & GSM8K & StrategyQA& &\\
    \midrule
    \multirow{5}{*}{Baichuan-7B} 
& Dense  & 0.00\% &19.1&9.84&55.5&28.1&100
    \\
    & LLMPruner & 24.2\% &12.6&\underline{1.74}&\underline{40.4}&18.2&64.8\\
    & Ours (None)  & 24.2\%&0.3&0&0&0.1&0 \\
    & Ours (FFN) & 25.1\% &\textbf{19.3}&\textbf{2.11}&\textbf{41.1}&\textbf{20.8}&\textbf{74.0}\\
    & Ours (Layer) & 24.2\% &\underline{18.2}&1.36&38.7&\underline{19.4}&\underline{69.0}\\
    \midrule
    \multirow{4}{*}{Baichuan-13B} 
    & Dense  & 0.00\% &24.6&27.1&61.1&37.6&100
    \\
    & Ours (None)  & 24.7\%&2.1&1.2&12.3&5.2&13.8 \\
    & Ours (FFN) & 25.5\% &\textbf{23.1}&\underline{2.1}&\textbf{47.3}&\textbf{24.2}&\textbf{64.4}\\
    & Ours (Layer) & 24.7\% &\underline{22.2}&\textbf{2.4}&\underline{43.2}&\underline{22.6}&\underline{60.1}\\

    \bottomrule
\end{tabular}
}
\label{tab:baichuan_gen}
\end{table}

\subsection{Experimental Results of Baichuan2-7B and Baichuan2-13B \label{sec: Result of Baichuan2-7B and Baichuan2-13B}}
We conduct experiments on Baichuan2-7B and Baichuan2-13B. The experimental results are shown in Table~\ref{tab:baichuan2-auc}, Table~\ref{tab:baichuan2_stability} and Table~\ref{tab:baichuan2_gen}.
The results indicate that our proposed LLM-Streamline is superior to the concurrent SOTA method, LaCo.

\begin{table}[H]
\caption{Accuracy of different pruning methods on classification benchmarks by pruning Baichuan2-7B and Baichuan2-13B. ``*'' indicates that we refer to the results in the original paper.}
	\small
	\centering
\renewcommand\arraystretch{1.2}
\resizebox{\linewidth}{!}{

	\begin{tabular}{c|c|c|cccccccccccc|cc}
		\toprule
		\multirow{2}{*}{LLM} & \multirow{2}{*}{Method}  & \multirow{2}{*}{Ratio}& \multicolumn{12}{c|}{Benchmarks}& \multirow{2}{*}{Average} & \multirow{2}{*}{RP} \\  
		&   & & C3 & CMNLI&CHID&BoolQ&WSC&CoQA&HeSW&PIQA&Race-M&Race-H &MMLU & CMMLU& &\\
		\midrule
		\multirow{6}{*}{Baichuan2-7B} 
  & Dense  & 0.00\% &64.4&33.4&85.5&63.1&42.3&63.1&67.6&76.1&51.1&52.5&54.7&57.1&59.2&100
		\\
		& LLMPruner & 24.2\% &39.9&\underline{33.9}&70.6&50.0&\textbf{42.3}&38.7&\underline{52.7}&\textbf{70.4}&22.3&22.8&24.9&24.9&41.1&69.4\\
		& LaCo*  & 24.2\%&50.9&33.0&\textbf{76.2}&56.2&\textbf{42.3}&47.3&52.3&\underline{68.5}&27.7&29.0&31.5&31.2&45.5&76.9\\
        & Ours (None)  & 24.2\%&45.7&33.0&58.0&\underline{62.6}&\underline{36.5}&41.9&46.3&62.4&25.6&27.4&43.0&46.5&44.1&74.5\\
        & Ours (FFN)  & 25.1\%&\underline{58.2}&33.0&\underline{74.1}&61.2&\underline{36.5}&\underline{47.6}&\textbf{54.3}&68.0&\underline{29.1}&\underline{30.5}&\underline{52.1}&\textbf{56.7}&\underline{50.1}&\underline{84.6}\\
        & Ours (Layer) & 24.2\%&\textbf{60.4}&\textbf{34.9}&72.2&\textbf{62.7}&\underline{36.5}&\textbf{48.8}&52.5&67.0&\textbf{35.5}&\textbf{36.8}&\textbf{54.0}&\underline{56.3}&\textbf{51.5}&\textbf{87.0}\\
            
            \midrule
            \multirow{5}{*}{Baichuan2-13B} 
  & Dense  & 0.00\% &65.6&33.2&86.7&66.8&42.3&65.6&71.1&78.1&68.9&67.2&59.6&61.3&63.9&100
		\\
		%& LLMPruner & 24.3\% &&&&&&&&&&&&&&\\
		& LaCo*  & 24.7\%&61.1&\underline{33.0}&76.7&\textbf{62.4}&\textbf{44.2}&\underline{55.5}&60.7&68.9&57.8&56.9&51.4&53.7&56.9&89.0\\
        & Ours (None)  &24.7\% &59.1&\textbf{34.4}&\underline{81.9}&61.8&36.5&53.9&61.9&\underline{71.0}&63.0&60.4&50.3&57.9&57.7&90.3\\
        & Ours (FFN)  & 25.5\%&\underline{63.0}&\underline{33.0}&81.7&60.1&36.5&54.7&\underline{62.1}&70.5&\textbf{71.1}&\textbf{68.2}&\underline{57.1}&\underline{58.2}&\underline{59.7}&\underline{93.4}\\
        & Ours (Layer) & 24.7\%&\textbf{63.5}&\underline{33.0}&\textbf{84.1}&\underline{62.0}&\underline{38.5}&\textbf{56.9}&\textbf{63.0}&\textbf{72.0}&\underline{70.2}&\underline{66.3}&\textbf{59.1}&\textbf{60.2}&\textbf{60.7}&\textbf{95.0}\\         
		\bottomrule
	\end{tabular}
 }
	\label{tab:baichuan2-auc}
\end{table}

\begin{table}[H]
\caption{Stability of different pruning methods on classification benchmarks by pruning Baichuan2-7B and Baichuan2-13B.}
	\small
	\centering
\renewcommand\arraystretch{1.2}
\resizebox{\linewidth}{!}{

	\begin{tabular}{c|c|c|cccccccccccc|c}
		\toprule
		\multirow{2}{*}{LLM} & \multirow{2}{*}{Method}  & \multirow{2}{*}{Ratio}& \multicolumn{12}{c|}{Benchmarks}& \multirow{2}{*}{Average} \\  
		&   & & C3 & CMNLI&CHID&BoolQ&WSC&CoQA&HeSW&PIQA&Race-M&Race-H &MMLU & CMMLU& \\
		\midrule
		\multirow{4}{*}{Baichuan2-7B} 
    		& LLMPruner & 24.2\% &62.2&50.6&\underline{75.1}&55.5&\underline{63.5}&66.0&\textbf{80.3}&\textbf{87.0}&54.2&51.7&50.0&46.9&61.9 \\
            & Ours (None) & 24.2\% &68.5&\textbf{98.1}&63.8&\textbf{69.1}&\textbf{84.6}&61.3&67.3&72.0&50.6&48.0&69.3&74.5&68.9\\
            & Ours (FFN) & 25.1\% &\underline{81.1}&\textbf{98.1}&\textbf{77.1}&\underline{68.2}&\textbf{84.6}&\underline{66.5}&\underline{78.1}&77.3&\underline{61.2}&\underline{57.7}&\underline{87.3}&\textbf{89.2}&\underline{77.2} \\
            & Ours (Layer) & 24.2\% &\textbf{83.3}&\underline{92.8}&74.9&67.6&\textbf{84.6}&\textbf{68.0}&77.2&\underline{80.9}&\textbf{64.1}&\textbf{62.9}&\textbf{89.2}&\underline{88.6}&\textbf{77.8} \\
            
            \midrule
            \multirow{3}{*}{Baichuan2-13B} 
            %& LLMPruner & 24.3\% &&&&&&&&&&&&& \\
            & Ours (None) & 24.7\% &85.0&\underline{88.7}&86.5&\underline{85.1}&\underline{82.7}&\underline{79.5}&82.2&85.1&\underline{84.6}&\underline{83.0}&74.2&85.6&83.5\\
            & Ours (FFN) & 25.5\% &\underline{86.4}&\textbf{99.0}&\underline{87.2}&84.7&\underline{82.7}&77.6&\underline{83.2}&\underline{85.7}&83.2&81.1&\underline{90.2}&\underline{91.7}&\underline{86.1} \\
            & Ours (Layer) & 24.7\% &\textbf{87.9}&\textbf{99.0}&\textbf{89.0}&\textbf{87.1}&\textbf{84.7}&\textbf{80.2}&\textbf{84.9}&\textbf{86.9}&\textbf{89.4}&\textbf{87.0}&\textbf{91.7}&\textbf{92.5}&\textbf{88.4} \\
            \bottomrule
	\end{tabular}
 }
	\label{tab:baichuan2_stability}
\end{table}

\begin{table}[H]
\caption{Accuracy of different pruning methods on generation benchmarks by pruning Baichuan2-7B and Baichuan2-13B. ``*'' indicates that we refer to the results in the original paper.}
\small
\centering
\renewcommand\arraystretch{1.2}
\resizebox{0.65\linewidth}{!}{
\begin{tabular}{c|c|c|ccc|cc}
    \toprule

    \multirow{2}{*}{LLM} & \multirow{2}{*}{Method}  & \multirow{2}{*}{Ratio}& \multicolumn{3}{c|}{Benchmarks}& \multirow{2}{*}{Average} & \multirow{2}{*}{RP} \\  
    &   & & Xsum & GSM8K & StrategyQA& &\\
    \midrule
    \multirow{6}{*}{Baichuan2-7B} 
& Dense  & 0.00\% &21.0&24.8&60.0&35.3&100
    \\
    & LLMPruner & 24.2\% &14.5&1.4&10.8&8.9&25.2 \\
    & LaCo* & 24.2\% &12.0&-&-&-&- \\
    & Ours (None) & 24.2\% &12.1&1.7&30.7&14.8&41.9\\
    & Ours (FFN) & 25.1\% &\underline{15.9}&\textbf{2.7}&\textbf{37.1}&\textbf{18.6}&\textbf{52.7}\\
    & Ours (Layer) & 24.2\% &\textbf{16.8}&\underline{2.3}&\underline{34.8}&\underline{18.0}&\underline{51.0}\\
    \midrule
    \multirow{5}{*}{Baichuan2-13B} 
    & Dense  & 0.00\% &25.3&53.2&65.9&48.1&100
    \\
   % & LLMPruner & 24.3\% &&&&& \\
    & LaCo* & 24.7\% &12.3&-&-&-&- \\
    & Ours (None)  & 24.7\%&17.2&\underline{3.3}&37.2&19.2&39.9 \\
    & Ours (FFN) & 25.5\% &\textbf{21.3}&3.1&\underline{48.8}&\underline{24.4}&\underline{50.7}\\
    & Ours (Layer) & 24.7\% &\underline{20.9}&\textbf{5.5}&\textbf{51.3}&\textbf{25.9}&\textbf{53.8}\\

    \bottomrule
\end{tabular}
}
\label{tab:baichuan2_gen}
\end{table}

\subsection{Experimental Results of Llama3.1-8B and Llama3.1-70B \label{sec: Llama3.1}}
We conduct experiments on Llama3.1-8B and Llama3.1-70B. The experimental results are shown in Table~\ref{tab:Llama3.1-auc} and Table~\ref{tab:Llama3.1_stability}.
The results indicate that our proposed LLM-Streamline is superior to the previous SOTA method.

\begin{table}[H]
\caption{Accuracy of different pruning methods on classification benchmarks by pruning Llama3.1-8B and Llama3.1-70B.}
	\small
	\centering
\renewcommand\arraystretch{1.2}
\resizebox{\linewidth}{!}{

	\begin{tabular}{c|c|c|cccccccccccc|cc}
		\toprule
		\multirow{2}{*}{LLM} & \multirow{2}{*}{Method}  & \multirow{2}{*}{Ratio}& \multicolumn{12}{c|}{Benchmarks}& \multirow{2}{*}{Average} & \multirow{2}{*}{RP} \\  
		&   & & C3 & CMNLI&CHID&BoolQ&WSC&CoQA&HeSW&PIQA&Race-M&Race-H &MMLU & CMMLU& &\\
		\midrule
		\multirow{4}{*}{Llama3.1-8B} 
  & Dense  & 0.00\% &65.3&33.0&73.8&68.2&36.5&69.8&74.7&81.1&71.6&64.5&66.8&52.5&63.2&100
		\\
        & SliceGPT  & 23.9\%&38.4&32.1&\underline{21.3}&37.8&\textbf{38.5}&\underline{38.0}&\underline{39.9}&58.7&21.9&23.3&25.8&25.2&33.4 &52.8\\
        & Ours (None)  & 24.4\%&\underline{42.3}&\underline{33.7}&19.3&\underline{52.3}&\underline{36.5}&30.7&28.4&\underline{58.9}&\underline{36.6}&\underline{33.3}&\underline{39.1}&\underline{34.4}&\underline{36.9}&\underline{58.4}\\
        & Ours (Layer) & 24.4\%&\textbf{55.9}&\textbf{34.5}&\textbf{54.5}&\textbf{67.6}&\underline{36.5}&\textbf{62.5}&\textbf{62.6}&\textbf{74.5}&\textbf{64.8}&\textbf{55.9}&\textbf{64.9}&\textbf{51.5}&\textbf{57.1}&\textbf{90.6}\\
        \midrule
        \multirow{4}{*}{Llama3.1-70B} 
        & Dense  & 0.00\% &74.8&33.0&81.6&76.5&37.5&73.0&79.9&83.9&86.8&80.5&79.3&68.8&71.3&100
		\\
        & SliceGPT  & 29.1\%&40.4&31.9&18.9&37.8&37.5&41.0&45.3&61.0&24.1&24.8&37.5&30.5&35.9&50.4\\
        & Ours (None)  & 30.3\%&\underline{66.1}&\textbf{37.5}&\underline{58.1}&\underline{69.0}&\textbf{46.2}&\underline{61.8}&\underline{68.4}&\underline{75.7}&\underline{81.7}&\underline{73.2}&\underline{70.4}&\underline{62.0}&\underline{64.2}&\underline{90.0}\\
        & Ours (Layer) & 30.3\%&\textbf{68.9}&\underline{34.7}&\textbf{70.0}&\textbf{72.5}&\underline{42.3}&\textbf{68.9}&\textbf{74.4}&\textbf{79.3}&\textbf{86.8}&\textbf{81.5}&\textbf{78.6}&\textbf{68.2}&\textbf{68.8}&\textbf{96.5}\\
                
		\bottomrule
	\end{tabular}
 }
	\label{tab:Llama3.1-auc}
\end{table}

\begin{table}[H]
\caption{Stability of different pruning methods on classification benchmarks by pruning Llama3.1-8B and Llama3.1-70B.}
	\small
	\centering
\renewcommand\arraystretch{1.2}
\resizebox{\linewidth}{!}{

	\begin{tabular}{c|c|c|cccccccccccc|c}
		\toprule
		\multirow{2}{*}{LLM} & \multirow{2}{*}{Method}  & \multirow{2}{*}{Ratio}& \multicolumn{12}{c|}{Benchmarks}& \multirow{2}{*}{Average} \\  
		&   & & C3 & CMNLI&CHID&BoolQ&WSC&CoQA&HeSW&PIQA&Race-M&Race-H &MMLU & CMMLU& \\
		\midrule
		\multirow{3}{*}{Llama3.1-8B} 
    		  & SliceGPT  & 23.9\%&63.2&37.0&38.5&35.6&\underline{98.1}&\underline{60.7}&\underline{59.3}&\underline{71.2}&39.0&46.1&42.1&47.4&53.2 \\
        & Ours (None)  & 24.4\%&\underline{59.9}&\underline{47.0}&\underline{39.0}&\underline{57.0}&\textbf{100}&49.8&43.4&64.7&\underline{53.4}&\underline{54.5}&\underline{57.8}&\underline{61.4}&\underline{53.7}\\
        & Ours (Layer) & 24.4\%&\textbf{78.5}&\textbf{49.7}&\textbf{59.9}&\textbf{75.0}&\textbf{100}&\textbf{80.3}&\textbf{84.8}&\textbf{86.5}&\textbf{87.3}&\textbf{86.1}&\textbf{90.8}&\textbf{89.1}&\textbf{80.7}\\
        \midrule
		\multirow{3}{*}{Llama3.1-70B} 
    		  & SliceGPT  & 29.1\%&55.6&\underline{45.4}&32.1&36.1&\textbf{98.1}&58.8&59.6&72.0&32.9&39.6&48.7&45.9&49.3 \\
        & Ours (None)  & 30.3\%&\underline{77.6}&43.6&\underline{64.5}&\underline{73.0}&91.4&\underline{76.8}&\underline{84.6}&\underline{85.0}&\underline{92.4}&\underline{89.3}&\underline{84.1}&\underline{81.4}&\underline{78.6}\\
        & Ours (Layer) & 30.3\%&\textbf{86.7}&\textbf{95.7}&\textbf{76.1}&\textbf{77.7}&\underline{95.2}&\textbf{89.5}&\textbf{92.6}&\textbf{93.4}&\textbf{97.2}&\textbf{96.3}&\textbf{95.9}&\textbf{94.6}&\textbf{90.9}\\
            
            \bottomrule
	\end{tabular}
 }
	\label{tab:Llama3.1_stability}
\end{table}

\subsection{Experimental Results of Mixtral-8x7B-v0.1 \label{sec: MOE}}
We conduct experiments on Mixture of Experts(MoE) model Mixtral-8x7B-v0.1. The experimental results are shown in Table~\ref{tab:MoE-auc} and Table~\ref{tab:MoE_stability}. 

\begin{table}[H]
\caption{Accuracy of LLM-Streamline on classification benchmarks by pruning Mixtral-8x7B-v0.1.}
	\small
	\centering
\renewcommand\arraystretch{1.2}
\resizebox{\linewidth}{!}{

	\begin{tabular}{c|c|c|cccccccccccc|cc}
		\toprule
		\multirow{2}{*}{LLM} & \multirow{2}{*}{Method}  & \multirow{2}{*}{Ratio}& \multicolumn{12}{c|}{Benchmarks}& \multirow{2}{*}{Average} & \multirow{2}{*}{RP} \\  
		&   & & C3 & CMNLI&CHID&BoolQ&WSC&CoQA&HeSW&PIQA&Race-M&Race-H &MMLU & CMMLU& &\\
		\midrule
		\multirow{3}{*}{Mixtral-8x7B-v0.1} 
  & Dense  & 0.00\% &54.1&33.0&48.6&68.4&56.7&68.2&76.2&81.7&72.4&70.9&71.3&52.8&62.9&100.0
		\\
        & Ours (None)  & 24.9\% &39.0&33.0&26.9&62.8&39.4&45.6&55.4&70.2&41.3&43.3&67.7&39.2&47.0&74.7\\
        & Ours (Layer) & 24.9\% &\textbf{51.2}&\textbf{34.4}&\textbf{41.5}&\textbf{66.3}&\textbf{56.7}&\textbf{62.0}&\textbf{68.3}&\textbf{77.9}&\textbf{54.5}&\textbf{55.7}&\textbf{69.9}&\textbf{50.2}&\textbf{57.4}&\textbf{91.3}\\
                
		\bottomrule
	\end{tabular}
 }
	\label{tab:MoE-auc}
\end{table}

\begin{table}[H]
\caption{Stability of LLM-Streamline on classification benchmarks by pruning Mixtral-8x7B-v0.1.}
	\small
	\centering
\renewcommand\arraystretch{1.2}
\resizebox{\linewidth}{!}{

	\begin{tabular}{c|c|c|cccccccccccc|c}
		\toprule
		\multirow{2}{*}{LLM} & \multirow{2}{*}{Method}  & \multirow{2}{*}{Ratio}& \multicolumn{12}{c|}{Benchmarks}& \multirow{2}{*}{Average} \\  
		&   & & C3 & CMNLI&CHID&BoolQ&WSC&CoQA&HeSW&PIQA&Race-M&Race-H &MMLU & CMMLU& \\
		\midrule
		\multirow{2}{*}{Mixtral-8x7B-v0.1} 
        & Ours (None)  & 24.9\% &66.9&\textbf{98.9}&62.4&76.4&38.5&69.3&72.4&80.6&57.0&60.3&79.1&64.4&68.9\\
        & Ours (Layer) & 24.9\%&\textbf{83.6}&87.6&\textbf{76.2}&\textbf{81.3}&\textbf{100}&\textbf{86.8}&\textbf{90.4}&\textbf{90.8}&\textbf{67.8}&\textbf{69.0}&\textbf{85.3}&\textbf{80.4}&\textbf{83.3} \\          
            \bottomrule
	\end{tabular}
 }
	\label{tab:MoE_stability}
\end{table}

\subsection{Experimental Results of OPT-1.3B and OPT-2.7B \label{sec: Result of OPT-1.3B and OPT-2.7B}}
We also conduct experiments on small models (OPT-1.3B and OPT-2.7B). The experimental results are shown in Table~\ref{tab:llm_comparison_all_OPT}.
The results indicate that our proposed LLM-Streamline is superior to the previous SOTA method, across different pruning rates.

\begin{table}[H]
\caption{Accuracy of different pruning methods by pruning OPT-1.3B and OPT-2.7B.}

\small
\centering
\renewcommand\arraystretch{1.2}
\resizebox{\linewidth}{!}{
\begin{tabular}{c|c|c|cccccc|cc}
    \toprule
    \multirow{2}{*}{LLM} & \multirow{2}{*}{Method}  & \multirow{2}{*}{Ratio}& \multicolumn{6}{c|}{Benchmarks}& \multirow{2}{*}{Average} & \multirow{2}{*}{RP} \\  
    &   & & PIQA & WinoGrande&HellaSwag&ARC-easy&ARC-challenge&OpenBookQA& &\\
    \midrule
    \multirow{12}{*}{OPT-1.3B} 
        & Dense  & 0.00\% &72.4 &59.3 &53.7  &51.0 &29.5  &23.4  &48.2 &100.0 \\ 
    & SliceGPT & 18.1\% & \underline{67.6} &\underline{53.6}& \underline{35.7} &\underline{51.1} &\underline{23.1}  &\underline{20.2}  &\underline{41.9}&\underline{86.9}\\
    & Ours(None) & 19.4\% &57.2 &51.7 &29.1 &32.5 &22.7 &13.2 &34.4 &71.4\\
    & Ours(FFN) & 18.1\% & \textbf{68.8} &\textbf{58.4}& \textbf{39.1} &\textbf{54.3} &\textbf{23.3}  &\textbf{23.3}  &\textbf{44.5}&\textbf{92.3}\\ 
        \cline{2-11} 
    & Dense  & 0.00\% &72.4 &59.3 &53.7  &51.0 &29.5  &23.4  &48.2 &100.0\\
    
    & SliceGPT & 25.8\% & \underline{65.5} &\underline{52.8}& \underline{34.2}&\underline{48.8}&\textbf{24.4}  &\underline{17.0}  &\underline{40.5}&\underline{84.0}\\
    & Ours(None) & 27.1\% &52.2 &51.1 &25.7 &26.6 &20.5 &14.0 &31.7 &65.8\\
    & Ours(FFN) & 25.8\% &\textbf{ 66.4} &\textbf{56.0}& \textbf{36.8} &\textbf{51.6} &\underline{22.2}  &\textbf{21.0}  &\textbf{42.3}&\textbf{87.8}\\ 
        \cline{2-11}
    & Dense  & 0.00\% &72.4 &59.3 &53.7  &51.0 &29.5  &23.4  &48.2 &100.0\\
    
    & SliceGPT & 33.6\% & \underline{62.4} &\textbf{52.6}& \underline{32.2}&\underline{45.4}&\textbf{23.1}  &\underline{16.6}  &\underline{38.7}&\underline{80.3}\\
    & Ours(None) & 34.8\% &50.5 &51.5 &25.8 &26.2 &20.3 &14.6 &31.5 &65.4\\
    & Ours(FFN) & 33.6\% & \textbf{62.9} &\underline{52.1}& \textbf{33.9} &\textbf{48.3} &\underline{20.8}  &\textbf{20.6}  &\textbf{39.8}&\textbf{82.6}\\ 
    \midrule
        \multirow{12}{*}{OPT-2.7B} 
        & Dense  & 0.00\% &73.8 &61.0 &45.9  &60.9 &26.8  &25.0  &48.9 &100.0\\
    & SliceGPT & 16.8\% &\underline{69.6} &\underline{56.3}&\underline{40.4} &\underline{56.2}&\textbf{27.5}  &\underline{20.2}  &\underline{45.0}&\underline{92.0}\\
    & Ours(None) & 17.8\% &61.2 &54.1 &33.8&41.2&24.1&15.8&38.4&78.5\\ 
    & Ours(FFN) & 16.8\% & \textbf{70.7} &\textbf{60.4}& \textbf{42.9} &\textbf{57.8} &\underline{25.3}  &\textbf{24.4}  &\textbf{46.9}&\textbf{95.9}\\ 
        \cline{2-11}
    & Dense  & 0.00\% &73.8 &61.0 &45.9  &60.9 &26.8  &25.0  &48.9 &100.0\\
    & SliceGPT & 25.7\% & \textbf{69.1} &\underline{55.0}& \underline{37.9}&\underline{53.9}&\textbf{26.7}  &\underline{18.2}  &\underline{43.5}&\underline{89.0}\\
    & Ours(None) & 26.7\% &59.7 &53.4&33.5&38.1&24.3&15.4&37.4&76.5\\ 
    & Ours(FFN) & 25.7\% & \underline{67.0} &\textbf{59.5}& \textbf{40.3} &\textbf{54.6} &\underline{24.7} &\textbf{22.2}  &\textbf{44.7}&\textbf{91.4}\\ 
        \cline{2-11}
    & Dense  & 0.00\% &73.8 &61.0 &45.9  &60.9 &26.8  &25.0  &48.9 &100.0\\
    & SliceGPT & 34.6\% & \underline{64.8} &\underline{54.1}&\underline{ 35.6}&\underline{50.0}&\textbf{26.5}  &\underline{18.0}  &\underline{41.5}&\underline{84.9}\\
    & Ours(None) & 35.6\% &56.6 &52.9 &31.5&37.6&24.1&14.9&36.3&74.2\\ 
    & Ours(FFN) & 34.6\% & \textbf{65.3} &\textbf{55.3}& \textbf{36.3} &\textbf{51.4} &\underline{24.5}  &\textbf{21.0}  &\textbf{42.3}&\textbf{86.5}\\ 
    \bottomrule
\end{tabular}
}
\label{tab:llm_comparison_all_OPT}

\end{table}

\subsection{Experimental Results of Llama2-7B at around 50\% pruning Ratio \label{sec:0.5}}
We conduct experiments on Llama2-7B at a higher pruning ratio of approximately 50\%. The experimental results are shown in Table~\ref{tab:Llama2-0.5-auc} and Table~\ref{tab:Llama2-0.5_stability}.
The results indicate that our proposed LLM-Streamline is superior to the previous SOTA method.

\begin{table}[H]
\caption{Accuracy of different pruning methods on classification benchmarks by pruning Llama2-7B at a higher pruning ratio of approximately 50\%.}
	\small
	\centering
\renewcommand\arraystretch{1.2}
\resizebox{\linewidth}{!}{

	\begin{tabular}{c|c|c|cccccccccccc|cc}
		\toprule
		\multirow{2}{*}{LLM} & \multirow{2}{*}{Method}  & \multirow{2}{*}{Ratio}& \multicolumn{12}{c|}{Benchmarks}& \multirow{2}{*}{Average} & \multirow{2}{*}{RP} \\  
		&   & & C3 & CMNLI&CHID&BoolQ&WSC&CoQA&HeSW&PIQA&Race-M&Race-H &MMLU & CMMLU& &\\
		\midrule
		\multirow{5}{*}{Llama2-7B} 
  & Dense  & 0.00\% &43.8 &33.0 &41.6  &70.8 &37.5  &66.7 &71.3 &78.1  &33.1  &35.5&46.8& 31.8  &49.2 &100.0
		\\
		& LLMPruner & 49.2\%&26.4&\underline{33.2}&\underline{17.3}&43.5&\underline{38.5}&25.4&32.7&\underline{59.2}&22.3&21.7&23.4&24.8&30.7&63.4\\
		& SliceGPT  & 48.3\%&26.5&32.1&15.4&38.1&\textbf{42.3}&28.1&30.9&53.6&23.6&23.1&25.2&25.3&30.4&61.8\\
        & Ours (None)  & 48.0\%&\underline{33.1}&\textbf{34.0}&\underline{17.3}&\underline{55.4}&36.5&\underline{31.0}&\underline{34.3}&56.3&\textbf{26.8}&\textbf{27.2}&\textbf{34.9}&\underline{27.9}&\underline{34.6}&\underline{70.3}\\
        & Ours (Layer) & 48.0\%&\textbf{39.7}&33.0&\textbf{27.7}&\textbf{62.1}&36.5&\textbf{44.3}&\textbf{45.2}&\textbf{63.6}&\underline{24.6}&\underline{24.3}&\underline{33.2}&\textbf{28.1}&\textbf{38.5}&\textbf{78.3}\\
            
		\bottomrule
	\end{tabular}
 }
	\label{tab:Llama2-0.5-auc}
\end{table}

\begin{table}[H]
\caption{Stability of different pruning methods on classification benchmarks by pruning Llama2-7B at a higher pruning ratio of approximately 50\%.}
	\small
	\centering
\renewcommand\arraystretch{1.2}
\resizebox{\linewidth}{!}{

	\begin{tabular}{c|c|c|cccccccccccc|c}
		\toprule
		\multirow{2}{*}{LLM} & \multirow{2}{*}{Method}  & \multirow{2}{*}{Ratio}& \multicolumn{12}{c|}{Benchmarks}& \multirow{2}{*}{Average} \\  
		&   & & C3 & CMNLI&CHID&BoolQ&WSC&CoQA&HeSW&PIQA&Race-M&Race-H &MMLU & CMMLU& \\
		\midrule
		\multirow{4}{*}{Llama2-7B} 
    	& LLMPruner & 49.2\%&63.6&\underline{96.2}&60.8&33.8&\underline{87.5}&51.5&\underline{55.5}&\underline{70.9}&47.8&48.1&56.9&58.1&60.9 \\
            & SliceGPT & 48.3\% &48.0&34.9&54.7&19.0&83.7&\underline{57.7}&52.7&63.4&\textbf{71.2}&\textbf{61.5}&52.2&58.1&54.8\\
            & Ours (None) & 48.0\% &\underline{73.2}&63.0&\underline{62.2}&\underline{65.2}&\textbf{95.2}&52.2&52.5&62.6&48.2&52.5&\underline{61.4}&\underline{58.8}&\underline{62.3}\\
            & Ours (Layer) & 48.0\% &\textbf{80.8}&\textbf{100}&\textbf{67.7}&\textbf{79.5}&\textbf{95.2}&\textbf{65.3}&\textbf{66.1}&\textbf{76.0}&\underline{58.7}&\underline{57.3}&\textbf{63.8}&\textbf{59.4}&\textbf{72.5}\\
            
            \bottomrule
	\end{tabular}
 }
	\label{tab:Llama2-0.5_stability}
\end{table}

\subsection{Results of Sufficient Post-training \label{Sufficient Post-training}}
Following the method outlined in Section \ref{sec: layer replacement and lora}, We conduct experiments using the entire SlimPajama-6B for post-training, and the results are presented in Table~\ref{tab: training data volumes}. As shown, using the entire dataset resulted in a slight improvement, but at a significant computational cost, requiring 100 times the computational time.

\begin{table}[H]
\caption{Detailed accuracy results with different training data volumes.\label{tab: training data volumes}}
	\small
	\centering
\renewcommand\arraystretch{1.2}
\resizebox{\linewidth}{!}{

	\begin{tabular}{c|c|c|ccccccccccccccc|cc}
		\toprule
		\multirow{2}{*}{LLM} & \multirow{2}{*}{Method}  & \multirow{2}{*}{Training data size}& \multicolumn{15}{c|}{Benchmarks}& \multirow{2}{*}{Average} & \multirow{2}{*}{RP} \\  
		&   & & C3 & CMNLI&CHID&BoolQ&WSC&CoQA&HeSW&PIQA&Race-M&Race-H &MMLU & CMMLU &Xsum &GSM8k &StrategyQA & &\\
		\midrule
		\multirow{3}{*}{Llama2-7B} 
    & Dense  &- &43.8 & 33.0 & 41.6 & 70.8 & 37.5 & 66.7 & 71.3 & 78.1 & 33.1 & 35.5 & 46.8 & 31.8 &19.4&16.5&60.2& 45.7 & 100.0\\
            & Layer-First & 30k &\textbf{43.9} & \textbf{33.0} & 29.8 & \textbf{70.8} & 36.5 & 59.6 & 64.3 & 73.4 & \textbf{36.6} & \textbf{37.4} & 44.9 & 30.0 & \textbf{19.7} & 2.05 & 54.8 & 42.5 & 93.0\\
		& Layer-First & 5.49M &43.5 & \textbf{33.0} & \textbf{33.2} & 68.8 & \textbf{46.2} & \textbf{61.1} & \textbf{66.5} & \textbf{76.0} & 31.8 & 29.9 & \textbf{47.3} & \textbf{31.8} & 18.2 & \textbf{10.6} & \textbf{58.6} & \textbf{43.8} & \textbf{95.8}\\
     
		\bottomrule
	\end{tabular}
 }
\end{table}

\subsection{Detailed Results of Different Lightweight Networks\label{Detailed result of different lightweight models}}

The detailed results of accuracy and stability of different lightweight networks on different classification benchmarks are shown in Table~\ref{tab: lightweight models acc} and Table~\ref{tab: lightweight models stability}. We can observe that FFN achieves the best results. Meanwhile, for the Transformer Layer, inheritance of the pruned first layer yields the best results.

\begin{table}[H]
\caption{Detailed accuracy results of different lightweight networks on different classification benchmarks, where ``$\dag$'' indicates that the intermediate size of the added lightweight network is half that of the default LLM's intermediate size. \label{tab: lightweight models acc}}
	\small
	\centering
\renewcommand\arraystretch{1.2}
\resizebox{\linewidth}{!}{

	\begin{tabular}{c|c|c|cccccccccccc|cc}
		\toprule
		\multirow{2}{*}{LLM} & \multirow{2}{*}{Method}  & \multirow{2}{*}{Ratio}& \multicolumn{12}{c|}{Benchmarks}& \multirow{2}{*}{Average} & \multirow{2}{*}{RP} \\  
		&   & & C3 & CMNLI&CHID&BoolQ&WSC&CoQA&HeSW&PIQA&Race-M&Race-H &MMLU & CMMLU& &\\
		\midrule
		\multirow{9}{*}{Llama2-7B} 
  & Dense  & 0.00\% &43.8 &33.0 &41.6  &70.8 &37.5  &66.7 &71.3 &78.1  &33.1  &35.5&46.8& 31.8  &49.2 &100.0
		\\
            & Layer-Random & 24.0\% &42.5  &\textbf{33.0}&\underline{27.0}  &65.9 &36.5  &58.0 &58.8 &70.2 &35.0 &36.0&\underline{46.3}   &31.4    &45.1 &91.7\\
            & Layer-First & 24.0\% &\underline{43.3}  &\textbf{33.0}&24.1  &\underline{67.5} &36.5  &\underline{59.2} &61.1 &\underline{71.5} &34.8 &37.0&45.5   &29.4    &45.2 &91.9\\
		& Layer-Last & 24.0\% &\textbf{43.5}  &\textbf{33.0}&\textbf{29.0}  &64.5 &\textbf{41.4}  &56.8 &\textbf{61.5} &\textbf{71.6} &34.5 &35.0 &46.0   &30.8    &45.6 &92.7\\
		& Layer-Avg & 24.0\% &42.1  &\textbf{33.0}&26.7  &66.3 &36.5  &57.7 &59.4 &69.9 &34.3 &34.5&43.3   &28.5    &44.4 &90.2\\
		& FFN$^\dag$ & 26.0\% &40.6  &\textbf{33.0}&24.2  &\underline{67.5} &36.5  &58.4 &59.5 &71.4 &\textbf{41.9} &\textbf{41.4}&\underline{46.3}   &30.8    &\textbf{46.0} &\textbf{93.5}\\
            & FFN & 25.0\% &40.7  &\textbf{33.0}&22.8  &65.9 &\underline{38.5}  &\textbf{60.6} &\underline{61.2} &71.2 &\underline{38.0} &\underline{38.7}&\textbf{47.0}  &\textbf{31.7}    &\underline{45.8} &\underline{93.1}\\
            & SwiGLU$^\dag$ & 26.0\% &41.9  &\textbf{33.0}&24.3  &\textbf{68.5} &36.5  &55.8 &57.9 &69.6 &29.9 &33.3&43.4   &\underline{31.6}    &43.8 &89.0\\
            & SwiGLU & 25.0\% &40.9  &\textbf{33.0}&22.1  &67.0 &36.5  &56.9 &59.1 &70.0 &33.8 &35.0&45.6   &30.8    &44.2 &89.8\\
     
		\bottomrule
	\end{tabular}
 }
\end{table}

\begin{table}[H]
\caption{Detailed stability results of different lightweight networks on different classification benchmarks, where ``$\dag$'' indicates that the intermediate size of the added lightweight network is half that of the default LLM's intermediate size. \label{tab: lightweight models stability}}
	\small
	\centering
\renewcommand\arraystretch{1.2}
\resizebox{\linewidth}{!}{

	\begin{tabular}{c|c|c|cccccccccccc|c}
		\toprule
		\multirow{2}{*}{LLM} & \multirow{2}{*}{Method}  & \multirow{2}{*}{Ratio}& \multicolumn{12}{c|}{Benchmarks}& \multirow{2}{*}{Average} \\  
		&   & & C3 & CMNLI&CHID&BoolQ&WSC&CoQA&HeSW&PIQA&Race-M&Race-H &MMLU & CMMLU& \\
		\midrule
		\multirow{8}{*}{Llama2-7B} 
            & Layer-Random & 24.0\% &79.9  &\textbf{100}&68.2  &83.8 &\textbf{95.2}  &77.4 &82.0 &83.2 &69.9 &71.9&85.1   &77.4   &81.2 \\
            & Layer-First & 24.0\% &79.8  &\textbf{100}&64.4  &\textbf{86.3} &\textbf{95.2}  &\textbf{81.7} &\underline{85.3} &\textbf{85.6} &81.8 &79.0&82.4   &71.0   &82.7 \\
		& Layer-Last & 24.0\% &\textbf{81.0}  &\textbf{100}&\textbf{73.5}  &\underline{84.9} &82.7  &\underline{81.3} &\textbf{85.8}&\underline{85.4} &82.3 &76.3&83.0  &66.9   &81.9 \\
		& Layer-Avg & 24.0\% &\underline{80.7}  &\textbf{100}&\underline{68.5}  &84.4 &\textbf{95.2 } &78.9 &82.8 &82.5 &73.0 &70.7&74.6   &58.8   &79.2 \\
		& FFN$\dag$ & 26.0\% &79.9  &\textbf{100}&65.4  &82.1 &\textbf{95.2}  &78.7 &80.7 &81.7 &74.7 &70.3&84.9   &74.6   &80.7 \\
            & FFN & 25.0\% &79.8  &\textbf{100}&64.1  &83.1 &\underline{93.3}  &80.7 &84.7 &84.6 &85.1 &79.0&\underline{87.5}   &\textbf{82.5}   &\textbf{83.7} \\
            & SwiGLU$^\dag$ & 26.0\% &78.5  &\textbf{100}&64.5  &78.9 &\textbf{95.2}  &75.4 &80.9 &82.1 &\textbf{89.3} &\textbf{87.1}&80.7   &78.4   &82.6 \\
            & SwiGLU & 25.0\% &78.9  &\textbf{100}&63.5  &84.4 &\textbf{95.2}  &77.0 &82.2 &82.3 &\underline{85.7}&\underline{82.4}&\textbf{87.9}   &\underline{79.7}   &\underline{83.3} \\
     
		\bottomrule
	\end{tabular}
 }
\end{table}

\subsection{Detailed Results of Different Pruning Ratio \label{Detailed result of pruning ratio}}

The detailed results of accuracy and stability on LLMs under different pruning ratios on different classification benchmarks are shown in Table~\ref{tab: different pruning ratios acc} and Table~\ref{tab: different pruning ratios stability}. The experiment results show that the performance of the pruned model is linearly correlated with the number of parameters, demonstrating the effectiveness of our method.

\begin{table}[H]
\caption{Detailed accuracy results of different pruning ratios  on different classification benchmarks, where ``$\dag$'' indicates that the intermediate size of the added lightweight network is half that of the default LLM's intermediate size. \label{tab: different pruning ratios acc}}
	\small
	\centering
\renewcommand\arraystretch{1.2}
\resizebox{\linewidth}{!}{

	\begin{tabular}{c|c|c|cccccccccccc|cc}
		\toprule
		\multirow{2}{*}{LLM} & \multirow{2}{*}{Method}  & \multirow{2}{*}{Ratio}& \multicolumn{12}{c|}{Benchmarks}& \multirow{2}{*}{Average} & \multirow{2}{*}{RP} \\  
		&   & & C3 & CMNLI&CHID&BoolQ&WSC&CoQA&HeSW&PIQA&Race-M&Race-H &MMLU & CMMLU& &\\
		\midrule
		\multirow{5}{*}{Llama2-7B} 
  & Dense  & 0.00\% &43.8 &33.0 &41.6  &70.8 &37.5  &66.7 &71.3 &78.1  &33.1  &35.5&46.8& 31.8  &49.2 &100.0
		\\
            & FFN$\dag$ & 2.0\% &\textbf{43.6}&\textbf{33.0}&\textbf{41.0}&\textbf{71.5}&\underline{36.5}&\textbf{65.4}&\textbf{70.7}&\textbf{77.5}&36.1&\underline{38.1}&\underline{46.3}&\textbf{31.6}&\textbf{49.3}&\textbf{100.0} \\
            & FFN$\dag$ & 11.0\% &42.1&\textbf{33.0}&\underline{35.1}&69.0&\textbf{37.5}&\underline{63.4}&\underline{67.8}&\underline{75.6}&35.2&37.0&\textbf{46.5}&29.9&\underline{47.7}&\underline{97.0} \\
            & FFN$\dag$ & 20.0\% &\underline{42.3}&\textbf{33.0}&29.0&\underline{70.2}&\underline{36.5}&62.7&63.8&72.3&\underline{37.8}&37.7&45.7&28.8&46.7&94.9 \\
            & FFN$\dag$ & 26.0\% &40.6  &\textbf{33.0}&24.2  &67.5 &\underline{36.5}  &58.4 &59.5 &71.4 &\textbf{41.9} &\textbf{41.4}&\underline{46.3}   &\underline{30.8 }   &46.0 &93.5\\
            \midrule
     \multirow{1}{*}{TinyLlama-1.1B}
     & Dense  & 0.00\% &38.3&34.6&30.4&56.4&47.1&48.8&54.5&71.3&24.1&25.8&25.8&25.0&40.2&100.0 \\
    
     \multirow{1}{*}{OpenLlama-3B-v2}
     & Dense  & 0.00\% &43.0&33.0&31.1&60.6&37.5&58.7&65.3&77.0&25.1&26.9&27.0&25.3&42.5&100.0 \\
		\bottomrule
	\end{tabular}
 }
\end{table}

\begin{table}[H]
\caption{Detailed stability results of different pruning ratios  on different classification benchmarks, where ``$\dag$'' indicates that the intermediate size of the added lightweight network is half that of the default LLM's intermediate size. \label{tab: different pruning ratios stability}}
	\small
	\centering
\renewcommand\arraystretch{1.2}
\resizebox{\linewidth}{!}{

	\begin{tabular}{c|c|c|cccccccccccc|c}
		\toprule
		\multirow{2}{*}{LLM} & \multirow{2}{*}{Method}  & \multirow{2}{*}{Ratio}& \multicolumn{12}{c|}{Benchmarks}& \multirow{2}{*}{Average} \\  
		&   & & C3 & CMNLI&CHID&BoolQ&WSC&CoQA&HeSW&PIQA&Race-M&Race-H &MMLU & CMMLU& \\
		\midrule
		\multirow{4}{*}{Llama2-7B} 
            & FFN$\dag$ & 2.0\% &\textbf{95.8}&\textbf{100}&\textbf{93.2}&\underline{86.5}&\textbf{95.2}&\textbf{96.2}&\textbf{97.8}&\textbf{97.1}&\textbf{93.0}&\textbf{89.5}&\textbf{95.5}&\textbf{89.7}&\textbf{94.1} \\
            & FFN$\dag$ & 11.0\% &\underline{90.4}&\textbf{100}&\underline{82.5}&\textbf{91.4}&\underline{94.2}&\underline{90.2}&\underline{93.4}&\underline{92.7}&\underline{89.6}&\underline{82.7}&84.6&72.0&\underline{88.6} \\
            & FFN$\dag$ & 20.0\% &81.6&\textbf{100}&71.9&86.0&\textbf{95.2}&84.4&87.3&87.9&85.2&78.6&78.1&67.7&83.7 \\
            & FFN$\dag$ & 26.0\% &79.9  &\textbf{100}&65.4  &82.1 &\textbf{95.2}  &78.7 &80.7 &81.7 &74.7 &70.3&\underline{84.9}   &\underline{74.6}   &80.7 \\
		\bottomrule
	\end{tabular}
 }
\end{table}

\subsection{Detailed Results of Layer Replacement and LoRA \label{detail result of continual fine-tuning}}

The detailed results of accuracy and stability of different layer replacement strategies and LoRA on different benchmarks are shown in Table~\ref{tab: layer replacement and LoRA acc} and Table~\ref{tab: layer replacement and LoRA stability}. The results show that layer replacement outperforms LoRA in both accuracy and stability.

%The average accuracy of TinyLlama-1.1B, Llama2-7B, and the models with different pruning ratios are shown in Fig.~\ref{fig: different pruning ratios.}. 
\begin{table}[H]
\caption{Detailed accuracy results of layer replacement and LoRA  on different classification benchmarks, where ``$\dag$'' indicates that the intermediate size of the added lightweight network is half that of the default LLM's intermediate size.  \label{tab: layer replacement and LoRA acc}}
	\small
	\centering
\renewcommand\arraystretch{1.2}
\resizebox{\linewidth}{!}{

	\begin{tabular}{c|c|c|cccccccccccc|cc}
		\toprule
		\multirow{2}{*}{LLM} & \multirow{2}{*}{Method}  & \multirow{2}{*}{Ratio}& \multicolumn{12}{c|}{Benchmarks}& \multirow{2}{*}{Average} & \multirow{2}{*}{RP} \\  
		&   & & C3 & CMNLI&CHID&BoolQ&WSC&CoQA&HeSW&PIQA&Race-M&Race-H &MMLU & CMMLU& &\\
		\midrule
		\multirow{9}{*}{Llama2-7B} 
  & Dense  & 0.00\% &43.8 &33.0 &41.6  &70.8 &37.5  &66.7 &71.3 &78.1  &33.1  &35.5&46.8& 31.8  &49.2 &100.0
		\\
            & Layer-First & 24.0\% &43.9  &\textbf{33.0}&\underline{29.8}  &\textbf{70.8} &\textbf{36.5}  &\underline{59.6} &\textbf{64.3} &\underline{73.4} &36.6 &37.4&44.9   &30.0    &\underline{46.7} &\underline{94.9}\\
		& Layer-Last & 24.0\% &\textbf{44.9}  &\textbf{33.0}&29.4 &\underline{69.2} &\textbf{36.5}  &58.9 &\underline{63.5} &\textbf{74.1} &\underline{37.8} &\underline{37.8}&\textbf{46.5 }  &30.4    &\textbf{46.8} &\textbf{95.1}\\
		& Layer-Avg & 24.0\% &43.9  &\textbf{33.0}&\textbf{30.1}  &67.5 &\textbf{36.5}  &58.3 &62.5 &72.3 &36.6 &36.1&\underline{46.2}   &\textbf{31.9}    &46.2 &93.9\\
		& FFN$\dag$ & 26.0\% &41.6  &\textbf{33.0}&25.8  &62.6 &\textbf{36.5}  &58.9 &62.1 &72.3 &\textbf{41.9} &\textbf{40.2}&44.4   &30.5    &45.8 &93.1\\
            & FFN & 25.0\% &43.8  &\textbf{33.0}&27.0  &68.7 &\textbf{36.5}  &\textbf{60.7} &\underline{63.5} &72.4 &37.4 &35.4&45.3   &\underline{31.5}    &46.3 &94.1\\
            & SwiGLU$^\dag$ & 26.0\% &\underline{44.0}  &\textbf{33.0}&27.9  &61.2 &\textbf{36.5}  &57.2 &61.7 &71.2 &30.3 &32.9&45.0   &\textbf{31.9}    &44.4 &90.2\\
            & SwiGLU & 25.0\% &43.2  &\textbf{33.0}&27.1  &67.0 &\textbf{36.5}  &58.2 &62.1 &71.2 &35.1 &35.7&45.8   &30.8    &45.5 &92.5\\
            &LoRA &24.0\% &43.2&\textbf{33.0}&27.6&63.5&\textbf{36.5}&57.7&62.4&71.7&30.7&32.9&43.5&30.8&44.5&90.4\\
     
		\bottomrule
	\end{tabular}
 }
\end{table}

\begin{table}[H]
\caption{Detailed stability results of layer replacement and LoRA  on different classification benchmarks, where ``$\dag$'' indicates that the intermediate size of the added lightweight network is half that of the default LLM's intermediate size.  \label{tab: layer replacement and LoRA stability}}
	\small
	\centering
\renewcommand\arraystretch{1.2}
\resizebox{\linewidth}{!}{

	\begin{tabular}{c|c|c|cccccccccccc|c}
		\toprule
		\multirow{2}{*}{LLM} & \multirow{2}{*}{Method}  & \multirow{2}{*}{Ratio}& \multicolumn{12}{c|}{Benchmarks}& \multirow{2}{*}{Average} \\  
		&   & & C3 & CMNLI&CHID&BoolQ&WSC&CoQA&HeSW&PIQA&Race-M&Race-H &MMLU & CMMLU& \\
		\midrule
		\multirow{8}{*}{Llama2-7B} 
            & Layer-First & 24.0\% &\underline{82.7}  &\textbf{100} &\textbf{74.3} &\underline{84.7}  &\textbf{95.2} &\underline{85.7} &\textbf{89.1} &\underline{88.6} &84.1&82.2   &84.5   &77.2 &\textbf{85.7}\\
		& Layer-Last & 24.0\% &\textbf{83.2}  &\textbf{100} &\underline{73.8} &\textbf{87.9}  &\textbf{95.2} &\textbf{89.5} &85.8 &\textbf{88.7} &84.7&82.1   &83.1    &72.9 &\underline{85.6}\\
		& Layer-Avg & 24.0\% &81.0  &\textbf{100} &72.3 &67.0  &\textbf{95.2} &84.1 &87.4 &86.6 &87.3&82.2   &\textbf{90.5}    &73.7 &83.9\\
		& FFN$\dag$ & 26.0\% &82.0  &\textbf{100} &71.4 &75.4  &\textbf{95.2} &82.9 &86.6 &87.2 &79.0&75.3   &85.1    &81.0 &83.4\\
            & FFN & 25.0\% & 80.6 &\textbf{100} &72.0 &83.5  &\textbf{95.2} &85.4 &\underline{87.7}&87.2 &84.5&81.0   &85.4    &79.3 &85.2\\
            & SwiGLU$^\dag$ & 26.0\% &80.0  &\textbf{100} &71.4 &63.1  &\textbf{95.2} &80.5 &86.1 &85.1 &\textbf{90.4}&\textbf{87.7}   &89.2    &\underline{87.7} &84.7\\
            & SwiGLU & 25.0\% &81.6  &\textbf{100} &72.9 &67.3  &\textbf{95.2} &81.8 &86.5 &85.3 &\underline{87.7}&\underline{84.8}   &\underline{90.0}    &\underline{83.3} &84.7\\
            & LoRA &24.0\% &81.9 &\textbf{100}& 73.4&59.1&\textbf{95.2}&81.6&84.8&85.3&85.2&81.3&82.1&75.3&82.1\\
     
		\bottomrule
	\end{tabular}
 }
\end{table}

\section{Limitation\label{sec: limitation}}

Our method achieves SOTA results compared to existing pruning methods, but its performance still falls short of other commonly used model compression methods, e.g., quantization. Therefore, we plan to enhance the performance of our pruning method and explore combining it with other compression and inference acceleration techniques to make it more practical.

\end{document}